%% file: main_v2.tex
\pdfoutput=1

\documentclass[11pt]{article}

\usepackage[final]{acl}

\usepackage{times}
\usepackage{latexsym}

\usepackage[T1]{fontenc}

\usepackage[utf8]{inputenc}

\usepackage{microtype}

\usepackage{inconsolata}
\usepackage{graphicx}
\usepackage{svg}
\usepackage{booktabs}
\usepackage{enumitem}
\usepackage{caption}
\usepackage{amsmath}
\usepackage{amssymb}
\usepackage{soul}
\usepackage{adjustbox}
\usepackage{xspace}
\usepackage[small,compact]{titlesec}
\usepackage{orcidlink}

\newcommand{\FRorcid}{\orcidlink{0000-0002-1697-8586}}
\newcommand{\MEAorcid}{\orcidlink{0000-0002-4360-1362}}
\newcommand{\YHorcid}{\orcidlink{0000-0003-3948-5845}}
\newcommand{\MLorcid}{\orcidlink{0000-0001-5765-0416}}

\newcommand{\benchmark}[0]{RECV\xspace}
\newcommand{\benchmarklong}[0]{\textbf{R}easoning in \textbf{E}vidence-based \textbf{C}laim \textbf{V}erification\xspace}
\newcommand{\vitc}[0]{VitaminC\xspace}
\newcommand{\climate}[0]{CLIMATE-FEVER\xspace}
\newcommand{\pheme}[0]{PHEMEPlus\xspace}

\newcommand{\posdelta}[1]{\textcolor{teal}{#1}}
\newcommand{\negdelta}[1]{\textcolor{red}{#1}}
\newcommand{\alignempty}[0]{$^{~~~~~~~~~~}$}

\newcommand{\squishlist}{
   \begin{list}{$\bullet$}
      { \setlength{\itemsep}{0pt} 
        \setlength{\parsep}{0pt} 
        \setlength{\topsep}{0pt} 
        \setlength{\partopsep}{0pt} 
        \setlength{\leftmargin}{1.5em} 
        \setlength{\labelwidth}{1em} 
        \setlength{\labelsep}{0.5em} } 
}

\newcommand{\squishend}{\end{list}}
\title{Assessing the Reasoning Capabilities of LLMs in the context of Evidence-based Claim Verification}

\author{John Dougrez-Lewis\textsuperscript{1,*},
    Mahmud Elahi Akhter\textsuperscript{2,*}\MEAorcid,
    Federico Ruggeri\textsuperscript{3}\FRorcid, \\
    \textbf{Sebastian Löbbers}\textsuperscript{2},
    \textbf{Yulan He}\textsuperscript{4,5}\YHorcid,
    \textbf{Maria Liakata}\textsuperscript{2,5}\MLorcid \\
    \textsuperscript{1}University of Warwick, UK,
    \textsuperscript{2}Queen Mary University of London, UK \\
    \textsuperscript{3}University of Bologna, Italy,
    \textsuperscript{4}King’s College London, UK \\
    \textsuperscript{5}The Alan Turing Institute, UK \\
  \texttt{j.dougrez-lewis@warwick.ac.uk},
  \texttt{yulan.he@kcl.ac.uk},
  \texttt{federico.ruggeri6@unibo.it} \\
  \texttt{\{m.akhter, s.lobbers, m.liakata\}@qmul.ac.uk} \\
  }

\begin{document}
\maketitle 
\begin{abstract}
Although LLMs have shown great performance on Mathematics and Coding related reasoning tasks, the reasoning capabilities of LLMs regarding other forms of reasoning are still an open problem. 
Here, we examine the issue of reasoning from the perspective of claim verification. 
We propose a framework 
designed to break down any claim paired with evidence into atomic reasoning types that are necessary for verification. 
We use this framework to create \benchmark, the first claim verification benchmark, incorporating real-world claims, 
 to assess the deductive and abductive reasoning capabilities of LLMs.
The benchmark comprises of three datasets, covering reasoning problems of increasing complexity.
We evaluate three state-of-the-art proprietary LLMs under multiple prompt settings.
Our results show that while LLMs can address deductive reasoning problems, they consistently fail in cases of abductive reasoning.
Moreover, we observe that enhancing LLMs with rationale generation is not always beneficial.
Nonetheless, we find that generated rationales are semantically similar to those provided by humans, especially in deductive reasoning cases.
\end{abstract}

\section{Introduction}
\label{sec:introduction}

Large Language Models (LLMs) have shown remarkable proficiency in complex tasks where reasoning capabilities, such as logical deduction and semantic comparison, are paramount.
Notable examples include solving MBA exams~\cite{Terwiesch2023}, passing professional medical tests~\cite{Kung2023, nori2023capabilities}, performing quantitative reasoning~\cite{lewkowycz2022solving}, and communication games~\cite{fair2022,Xu2023ExploringLL,Gandhi2023StrategicRW}.
However, there is ongoing debate about whether such proficiency is due to LLMs manifesting reasoning capabilities or rather pattern matching and semantic similarity via memorization.
For example, earlier claims that LLMs posses Theory of Mind (ToM) capabilities~\cite{bubeck2023sparks,kosinski2023theory} were shown to be inaccurate~\cite{ullman2023large, sileo-lernould-2023-mindgames}.
In particular, despite appearing to manifest some form of ToM capabilities, LLMs 
mostly rely on shallow heuristics and spurious correlations~\cite{Shapira2023CleverHO}.
Additionally, preliminary observations of emergent reasoning capabilities~\cite{wei2022emergent} were subsequently attributed to metric choice~\cite{schaeffer2023emergent}, in-context learning~\cite{lu2023emergent}, and shortcuts~\cite{kavumba-etal-2019-choosing}.

These findings motivate the need for further research on the reasoning capabilities of LLMs, especially in high-stake real-world applications, where research on this topic is in its infancy.
A notable example is fact-checking, where LLMs are considered to hold great potential for increased productivity even if at the same time they also facilitate bad actors in the proliferation of misinformation~\cite{guo2023close} 
Verifying information is challenging since models require both accurate veracity classification and strong rationale generation to be effective~\cite{schlichtkrull-etal-2023-intended}.
It is thus essential to understand the reasoning capabilities and limitations of LLMs 
in the context of fact-checking.
In particular, we extend the current discussion around the reasoning abilities of LLMs, focusing on their ability to verify real-world claims. 


In this work, we first propose a framework for breaking down complex claims into atomic reasoning steps. 
The motivation behind this is the lack of uniform terminology around reasoning evaluation. 
Most prominent evaluation datasets for reasoning are based on mathematics and coding, which involve deductive reasoning, even though the use of deduction is not made explicit~\cite{{sprague2024cotcotchainofthoughthelps}}. 

Our framework is rooted in existing philosophy literature concerning logical reasoning that aligns well with NLP~\cite{Wason1972-WASPOR-3,Galotti1989ApproachesTS}.
We use our framework to create \benchmarklong (\benchmark), the first reasoning benchmark for claim-verification.
The benchmark comprises three datasets, curated from existing resources targeting different domains: \vitc~\cite{schuster-etal-2021-get} from Wikipedia, \climate~\cite{diggelmann2020climatefever} from online claims and Wikipedia, and \pheme~\cite{phemeplus} from rumours circulating on social media and associated evidence from news articles. 
The claims involve increasing levels of complexity as we move from \vitc to \pheme, often requiring deductive and/or abductive reasoning. 

We use \benchmark to evaluate three state-of-the-art proprietary LLMs that have shown impressive performance on various reasoning and language benchmarks~\cite{huang-chang-2023-towards, deepseekai2025deepseekr1incentivizingreasoningcapability}.
These models are: Claude V3 Sonnet~\cite{anthropic-2023-claude-v3}, GPT-4~\cite{openai-2024-gpt4-technical-report}, and GPT-4o~\cite{openai-2024-gpt4o-card}.
In particular, we prompt models with and without Chain-of-Thought (CoT)~\cite{wei2023chainofthought} rationale generation to assess if and how the latter influences reasoning. 
In alignment with previous work~\cite{saparov2023testing,akyürek2024deductive,li-etal-2024-look} we find that LLMs are capable of deductive reasoning.
However, they consistently fail at claim verification when presented with evidence that requires abductive reasoning.
Furthermore, we observe conflicting results when prompting LLMs with CoT strategies.
In particular, CoT leads to performance improvements for simple claim verification as in \vitc, 
but is detrimental in the case of complex claims such as those found in \climate and \pheme.
Lastly, we carry out a qualitative analysis of generated rationales and observe high semantic similarity with human explanations, especially in deductive reasoning cases.
In summary, we make the following contributions:



    

\squishlist
    \item We propose a framework for decomposing claim-evidence pairs into atomic reasoning types for verification, covering deductive and abductive reasoning (\S\ref{sec:methodology}).

    \item We create the first reasoning benchmark for claim verification comprising three datasets of increasing complexity (\S\ref{sec:benchmark}).

    \item We extensively evaluate the reasoning capabilities of three state-of-the-art LLMs, showing that models fail when it comes to abductive reasoning and CoT's effectiveness is task-dependent (\S\ref{sec:experiments}).

    \item We show that generated rationales are consistent with human reasoning for correct predictions, but model are often unable to leverage such rationales for claim verification (\S\ref{sec:qualitative-analysis}).
\squishend

\input{figures/framework}

\section{Terminology} \label{sec:terminology}

To avoid ambiguity, we provide a brief overview of the main terminology used in the paper.
We first introduce fact-checking and rumour verification specific terminology.
We define a \textbf{claim} as a check-worthy statement, i.e., a piece of information that has to be verified.
Similarly, a \textbf{rumour} is a widely circulating claim of unknown veracity~\cite{zubiaga-etal-2019-detection}.
An \textbf{evidence} is a piece of information, here textual, retrieved from a document, that can be used to verify the veracity of a claim.

We also provide an overview of reasoning-specific terminology.
We denote \textbf{conclusion} to be a statement that derives from the direct elaboration of some observations.
An \textbf{observation} is a piece of factual information.
In the case of a reasoning task for claim verification, \textbf{observations} coincide with \textbf{evidence}, as reasoning is performed on factual information with the intent of verifying a given claim.
Lastly, we denote \textbf{explanations} as natural language descriptions of the reasoning process that leads to a conclusion from a given set of observations.


\section{ RECV Logical Framework} \label{sec:background}

Reasoning is often used interchangeably to denote critical thinking, decision-making, and logical reasoning.
Following \citet{Wason1972-WASPOR-3} and \citet{Galotti1989ApproachesTS}, we define reasoning as the process of logical steps that result in some form of decision-making or conclusion. 
Thus we define reasoning as a series of inference steps linking claims and evidence to reach a conclusion.

In particular we consider that reasoning consists of the interplay of three interrelated components: \textit{types}, \textit{processes}, and \textit{tasks}. This is the basis of our RECV framework. 
Reasoning \textit{types} are different forms of logical inference that we can use to reach a conclusion from a set of observations or premises.
We distinguish between atomic and compound reasoning types.
Atomic types denote basic forms of logical inference and include deductive, abductive, inductive, and analogical reasoning.
A reasoning \textit{task} is any task that requires multiple reasoning types, often in complex interaction with each other. For example claim verification is a composite reasoning task. 
A reasoning \textit{process} is the method of interaction between reasoning types or even tasks within complex reasoning tasks. 
Notable examples of reasoning processes are multi-hop or multi-step inference, where individual steps or hops can be of different reasoning types.
In this paper we focus particularly on atomic reasoning types.

\subsection{Atomic Reasoning Types}


\paragraph{Deduction:}
A conclusion is drawn directly from evidence. In the context of a claim, if the evidence supports the claim then the claim is deduced to be true (if $P$ then $Q$, where $P$ is the claim and $Q$ is the evidence). For example, 

\squishlist
    \item $P$: \textit{Schools closed, Dammartin-en-Goele residents told to stay indoors, town `like warzone'.} \textbf{[Claim]}

    \item $Q$: \textit{Schools went into lockdown and the town appealed to residents to stay inside residents' houses.} \textbf{[Evidence]}

    \item $C$: Here, $P\implies Q$. The schools have been closed and citizens have been told to stay home. Thus, the town is like in a warzone situation. \textbf{[Conclusion]}
\squishend    

Equally if the evidence contradicts the claim then the claim is deduced to be false. 

\squishlist
    \item $P$: \textit{Heart goes out to 148 passengers and crew of Germanwings Airbus A320 that has crashed in French Alps, Southern France.} \textbf{[Claim]}

    \item $Q$: \textit{ German jetliner carrying 144 passengers and six crew en route from Barcelona, Spain, to Düsseldorf, Germany, has crashed in the French Alps, killing all 150 people on board.} \textbf{[Evidence]}

    \item $C$: Here, $Q$ contradicts $P$. The evidence directly states the death toll is 150 which refutes the claim. \textbf{[Conclusion]}
\squishend


\vspace{-0.2cm}
\paragraph{Abduction:} 
The most plausible conclusion is drawn from a set of candidate hypotheses, based on partial evidence.
Abduction could lead to false conclusions.

\squishlist
   \item \textbf{Claim:} \textit{ Pluto's climate change over the last 14 years is likely a seasonal event.}
   
   \item \textbf{Evidence:} \textit{ The long orbital period of Neptune results in seasons lasting forty years. As a result, Neptune experiences similar seasonal changes to Earth. There's evidence for methane escape and strong seasonal and dynamical perturbations of Neptune's atmospheric temperatures. Each planet therefore has seasons, changes to the climate over the course of its year.}
   
   \item \textbf{Conclusion:} \textit{The evidence only mentions Neptune. However, the claim is regarding Pluto. Given the partial evidence, the claim is supported based on the plausible hypothesis that Pluto is near Neptune and it is likely to have similar attributes when it comes to seasons and climate change.
   }
\squishend

\paragraph{Induction:} 
An inference is drawn from complete evidence (in a specific domain) and then a generalization (a rule that can be used beyond the initial domain) is derived from it. 
%
As per \citet{Flach2000}, for inductive reasoning, the evidence can be true whilst only providing partial support for the conclusion, which typically generalizes beyond the evidence itself. 
Such generalization indicates there is no guarantee that the conclusion is true elsewhere. 

\paragraph{Analogical reasoning:} 
Conclusions are drawn based on the similarities between entities.
%
While we do not provide examples for inductive and analogical reasoning in this section, they are still part of our framework. 
The focus on deduction and abduction is justified in (\S\ref{sec:methodology}).
We provide more formal definitions of atomic reasoning types in Appendix~\ref{app:reasoning} with additional examples. 

\section{Methodology} \label{sec:methodology}

We discuss our methodology for reasoning in claim verification.
We first showcase the application of our RECV framework and then motivate our focus on deduction and abduction via a preliminary study. 
\paragraph{RECV Logical Framework Application:}
The application of the RECV framework can be seen in Figure~\ref{fig:reasoning_path}. The \textit{reasoning task} here is claim verification and the \textit{reasoning type} is abductive.  
The claim is resolved using a \textit{single-step process}, that consists of abductive \textit{type atomic reasoning}. Here we only highlight the most plausible hypothesis that resolves the claim as \textit{true}. However, in practice, we would generate multiple hypotheses before coming to the most plausible one. 

\paragraph{Preliminary Study:}
Our objective here was to determine the atomic reasoning types necessary for accomplishing claim verification from evidence. 
We first collected a small dataset by manually selecting 90 claims and associated evidence from \vitc~\cite{schuster-etal-2021-get}, \climate~\cite{diggelmann2020climatefever}, and 
\pheme~\cite{phemeplus}. We focus on these resources as they are widely used in claim and rumour verification and differ in complexity.
Two annotators with expertise in Computer Science and native English proficiency assigned reasoning type labels to claim-evidence pairs 
following (\S\ref{sec:background}).
Disagreements encountered were resolved via a discussion stage with an independent expert.
The Inter-Annotator Agreement (IAA) measured as Bennett's S score~\cite{10.1086/266520} to account for label imbalance of reasoning types is 0.90, denoting almost perfect agreement.
We observe that all examples either require deductive or abductive reasoning types.

\paragraph{Deductive and Abductive Reasoning:}
\label{deduc_rationale}
Our preliminary investigation suggests that inductive and analogical reasoning are rarely employed in claim verification. 
This is presumably because inductive reasoning relies on complete evidence, which is rarely available in real-world domain-specific settings. 
Generalisations from one domain to another, relevant to inductive reasoning, may only occur in scenarios that share common background knowledge, as in the medical domain. 
Similarly, analogical reasoning may be more suitable for other fact-checking related tasks like profiling and motive analysis where frequent and repeated comparisons may occur to reach a conclusion. 
By contrast, deductive and abductive reasoning types are often required in fact-checking~\cite{pan-etal-2023-fact,tan-etal-2024-enhancing-fact}.
For these reasons, here we focus on deduction and abduction.
We show that they represent a challenging setting for claim verification (\S\ref{sec:benchmark}), 
and model evaluation with LLMs (\S\ref{sec:experiments}).

\section{\benchmark Benchmark} \label{sec:benchmark}

We discuss the creation of \benchmark, 
in particular, our sample selection strategy and data annotation.
See Appendix~\ref{app:datasets} for details regarding the three datasets. 

\paragraph{Data Sampling Strategy.}
We build a heuristic-based sampling strategy to mitigate the anticipated data imbalance between deductive and abductive samples, as it was important to ensure both are represented in the annotated data.
%
We used a combination of three embedding-based text similarity metrics to compute the average claim similarity between deductive samples.
Likewise for abductive samples.
The metrics are: cosine similarity (SimScore), BERTScore~\cite{bert-score} and BLEURT score~\cite{pu2021learning}.
We used the data collected during our preliminary study (\S\ref{sec:methodology}) to set a similarity threshold for each reasoning type.
In particular, we computed the distribution of similarity metrics in the preliminary study data. 
We computed a separate distribution for deductive and abductive samples, respectively. See Figure \ref{fig:sampline_threshold} in Appendix~\ref{sec:appx_sample} for a summary.
We found that abductive samples had lower BERTScore and higher lexical similarity (SimScore), likely due to indirect evidence, with low surcace overlap.
Deductive samples, by contrast, had higher BLEURT and SimScore.
From these observations, we derived two filtering thresholds.
For abductive samples, the threshold is: $\text{BERTScore} \leq 0.25 \,\wedge \, \text{SimScore} \geq 0.35$.
For deductive samples, the threshold is: $\text{SimScore} \geq 0.36 \,\wedge \, \text{BLEURT} > 0.15$.

We used each threshold to sample claims likely to be resolved via deductive and abductive reasoning, respectively.
In particular, we exclude instances 
labeled as `unverified' since such claims are always associated with deductive reasoning, either due to lack of appropriate evidence or contradictory evidence. 
In total, we sampled 500 claim-evidence pairs from each dataset.
This strategy allowed us to avoid an over-representation of deductive cases,ensuring diversity in the annotated data. 
However, we remark that this strategy is noisy at best and only provides a weak guarantee.
See Appendix~\ref{sec:appx_sample} for more details on data sampling.

\input{tables/benchmark_stats}


\paragraph{Data Annotation}
We recruited 9 PhD students in Computer Science, fluent in English and grouped them in triples, one for each dataset.
The annotation task involved labeling claim-evidence pairs as requiring either deductive or abductive reasoning in order to be resolved. More precisely, annotators labeled claim-evidence pairs with the reasoning types \textbf{deductive} or \textbf{abductive}. Table~\ref{fig:methodology} (top) in Appendix~\ref{app:guidelines} summarizes the annotation process.
We evenly distributed dataset samples to annotators in a triple, so that 100 samples were annotated by all.
In total, each annotator in a triple labeled 233 samples.
Annotation guidelines per dataset are in Appendix~\ref{app:guidelines}.
We computed IAA as Bennett's S score to account for label imbalance (see Appendix~\ref{app:data-annotation} for pairwise agreement scores).
The IAA is 0.75 for \vitc, 0.56 for \climate, and 0.67 for \pheme.
Table~\ref{tab:claim_distribution} reports our \benchmark statistics.
In particular, we observe that the rate of abductive reasoning samples is relatively low compared to deductive ones: 5.8\% in \vitc, 20.4\% in \climate, and 7.2\% in \pheme.
This imbalance is expected given the nature of collected evidence; most evidence provided, either in the form of Wikipedia articles as in \vitc and \climate or news articles as in \pheme, contains detailed information to deductively verify the claim. In total, \benchmark consists of 1500 claim-evidence pairs with associated veracity and reasoning labels. The average sentence length for evidence in \vitc is 1.084, 7.562 for \pheme and 7.828 for \climate. This highlights the varying complexity of the datasets and \benchmark. 


\section{Claim Verification with LLMs} \label{sec:experiments}
Our objective here is to assess the capabilities of LLMs in performing deductive and abductive reasoning to determine the veracity of a claim. 


\paragraph{Setup:}
We formulate claim verification as a prediction task. Given a claim-evidence pair, we prompt LLMs to predict whether the evidence supports or refutes the claim (Figure~\ref{fig:methodology} (bottom)). 
We consider two settings: \emph{\textbf{No-Exp}} and \emph{\textbf{Exp}}.
In \emph{\textbf{No-Exp}}, we prompt LLMs to predict the claim veracity without any rationale generation.
In \emph{\textbf{Exp}}, we first prompt LLMs to produce a rationale and then use the generated information to predict claim veracity.
For each setting, we consider two different prompt strategies: Zero-Shot (ZS), and Manual Chain-of-Thought (M-CoT)~\cite{{wei2023chainofthought}}.
In addition, in \emph{\textbf{Exp}}, we also consider Zero-Shot Chain-of-Thought (ZS CoT)~\cite{{kojima2023large}}. ZS CoT was applied only under \emph{\textbf{Exp}} as rationale generation is integral to ZS CoT prompting. In all the prompts, we provide dataset specific personas and instructions in the system prompt and CoT examples in the user prompt. We report the prompts in Appendix~\ref{app:prompts}.

\paragraph{Metrics:}
We compute macro F1 score for veracity of claims given the evidence and the error rate of claim-evidence pairs concerning deductive and abductive reasoning types, respectively. F1 was chosen due to the class imbalance in \climate and \pheme (they have a 70/30 ratio between support and refute labels). 
We use annotators' reasoning type labels for claim-evidence pairs to identify errors in verification per category (cases of abduction vs deduction) and express it via error rate.



\paragraph{Models:}
We consider three state-of-the-art proprietary LLMs with remarkable proficiency in a wide range of tasks: Claude V3 Sonnet~\cite{anthropic-2023-claude-v3}, GPT-4~\cite{openai-2024-gpt4-technical-report}, and GPT-4o~\cite{openai-2024-gpt4o-card}.  We conducted our experiments using OpenAI and Anthropic's official API.


\subsection{Results} \label{sec:results}

Table~\ref{tab:claim-verification} reports classification performance and error rates per reasoning type for claim verification on \benchmark.
We discuss dataset-specific results in detail.

\input{tables/merged_results}

\paragraph{\vitc:}
Among prompting strategies, M-CoT leads to the highest increase in performance across all models.
The average error rate across all models and settings is $10.31\%$ for deductive reasoning and $32\%$ for abductive reasoning.
\textit{This shows that all models struggle with abductive reasoning, even in less challenging settings like \vitc.}
Regarding model settings, we observe conflicting results.
In particular, generating rationales improves veracity classification performance for deductive samples in all models, except for GPT-4 M-CoT.
By contrast, only Claude ZS and GPT-4o ZS show improvements in \emph{Exp} compared to \emph{No-Exp} when targeting abductive reasoning.
Overall, when moving to the \emph{Exp} settings, we observe a $7.5\%$ average performance drop, with GPT-4 reporting the highest degradation ($-10\%$).
Lastly, regarding prompting strategies, we observe that M-CoT outperforms CoT in deductive cases, while reporting comparable results in abductive ones.

\paragraph{\climate:}
The average error rate across all models and settings is $15.58\%$ for deductive reasoning and $48.58\%$ for abductive reasoning.
Similar to \vitc, these results denote that LLMs fail at predicting claim veracity when dealing with abductive reasoning.
In particular, abductive reasoning samples are on average three times more challenging than deductive ones.
Regarding model settings, we observe that rationale generation leads to performance degradation in all scenarios.
Overall, we observe a $4.36\%$ average performance drop for deductive cases and $14.76\%$ for abductive ones.
Regarding prompting strategies, we observe similar results to \vitc where M-CoT outperforms CoT.
In particular, the average error rate for M-CoT is $14.24\%$ on deductive cases ($+2.76$) and $45.1\%$ on abductive ones ($+8.17$).

\paragraph{\pheme:}
The results suggest that there is no model or prompting strategy that consistently outperforms others. The average error rate across all models and settings is $20.06\%$ for deductive reasoning and $44.68\%$ for abductive reasoning.
Compared to \vitc and \climate, \pheme represents a more challenging setting for deductive reasoning, while it is comparable in complexity with \climate when assessing LLMs for claim verification.
Regarding model settings, we observe minor performance improvements when prompting LLMs to generate rationales in deductive cases, with a $1.46\%$ average gain.
Claude is the only exception with a $2.81\%$ average performance drop when moving to \emph{Exp}. 
By contrast, we observe notable performance degradation in abductive reasoning cases, with GPT-4 ZS \emph{Exp} being the worst ($-13.89\%$).
Lastly, regarding prompting strategies, we observe no performance difference between ZS CoT and M-CoT, highlighting the higher task complexity in \pheme.

\section{Explanation evaluation} \label{sec:qualitative-analysis}
Providing reasonable explanations to support predicted veracity labels is a crucial aspect of claim verification systems.
In particular, an automated system needs to be both convincing and trustworthy to convince users in practice~\cite{schlichtkrull-etal-2023-intended}.
Therefore, we evaluate the LLMs generated rationales in the \emph{Exp} setting.
This is crucial considering that LLMs tend to hallucinate~\cite{bouyamourn-2023-llms,rawte-etal-2023-troubling} and be self-contradictory at times~\cite{mündler2023selfcontradictory}.
We randomly selected 100 samples from each dataset in \benchmark and compared generated rationales against those provided by human annotators.
We note that the provided human rationales explain the reasons for the chosen veracity and do not explicitly have any mention of why the given rationale corresponds to a reasoning type.
In particular, the main goal of this study is to evaluate the models' reasoning capabilities through implicit measurement and without biasing the models with the mention of reasoning modes.
This is also equivalent to how models are prompted, where we do mention reasoning types.
We instruct a third annotator to evaluate the quality of provided human rationales. See Appendix \ref{app:human-rationale} for more details.
We restricted sample selection to those where at least three models predicted wrong veracity labels.
We follow \citet{song-etal-2024-combining} and compute \emph{Factual Consistency} (FC), \emph{Evidence Appropriateness} (EA), \emph{BARTScore}~\cite{NEURIPS2021_e4d2b6e6}, and \emph{Perplexity} (PPL) to assess the quality of the generated explanations.
We provide additional details about metrics in Appendix~\ref{app:qualitative analysis}.


\subsection{Results}

\input{tables/human_evaluation}

Table~\ref{tab:human-evaluation} reports the results concerning explanation evaluation.
Appendix \ref{app:statistical-significance} reports statistical significance results to support our findings.
We observe that all models achieve comparable results on appropriateness (EA), consistency (FC), and coherence measured via BARTScore (BART), while showing notable discrepancies regarding perplexity (PPL).
In particular, GPT-4o ZS CoT has the most faithful rationales across all datasets.
Moreover, prompting strategies like ZS CoT and M-CoT do not lead to consistent improvements over ZS, suggesting that their effectiveness may be problem- and model-dependent.

Additionally, we assess generated rationales regarding correct and wrong model predictions in Appendix~\ref{app:qualitative analysis}.
Our results show that rationales from correct predictions better align with ground-truth explanations, suggesting that wrong predictions are usually the by-product of incorrect reasoning (the model is unable to leverage the explanation).

Lastly, we analyze how similar the generated rationales were between the models.
To do so, we perform a permutation test using sentence-level contradiction scores from Fact\_Score (see Appendix~\ref{app:qualitative analysis}).
We find that Claude ZS has the most unique rationales on all datasets. 


We discuss properties of rationales generated in the case of abductive and deductive errors per dataset.



\paragraph{\vitc}
We observe that LLMs struggle to generate faithful rationales in abductive cases.
In particular, models tend to generate assertions rather than hedged information.
This has implications for claim verification, where models predominantly refute or misclassify the veracity of the claim based on the generated explanations.
Regarding deductive reasoning, we observe that the majority of errors are due to internal biases of LLMs, heavily influencing rationale generation, and to semantic faults in attending to only some parts of the claim and evidence.


\paragraph{\climate}
Regarding abductive cases, we observe the same issue reported in \vitc. 
Regarding deductive reasoning, the majority of failures are due to implicit reasoning where relevant evidence information is implicit or where temporal relations between factual content must be understood to reach the correct conclusion.


\paragraph{\pheme}
Contrary to \vitc and \climate, abductive and deductive reasoning errors are mainly due to semantic interpretation issues where models focus only on specific information in the claim and evidence.
This limits models in assessing claim-evidence pairs in their entirety, thus, hindering them in capturing relations between the evidence and the claim. 
As in \vitc, this issue affects the claim verification performance, often leading to misclassification.


\section{Findings} \label{sec:discussion}

We discuss the main findings of our work, including task complexity, the effectiveness of prompting strategies, and rationale generation.

\paragraph{Reasoning and Task complexity.} 
Our results show that abductive reasoning is consistently more challenging than deductive reasoning.
In particular, the performance gap between the two cases is around three times on average.
This is mainly motivated by LLMs failing in performing uncertainty reasoning, often leading to erroneous assertive conclusions.
Nonetheless, this is not the only issue that makes \benchmark challenging; task complexity plays a crucial role in reasoning performance.
For instance, \pheme represents a more complex setting than \vitc where news articles can contain extensive amount of information compared to Wikipedia pages.
As shown in our qualitative analysis, LLMs tend to focus only on specific parts of input claim-evidence pairs, leading to suboptimal performance. For example, in the following claim from \vitc,

\textit{Claim:} Peking University is a unitary sovereign state that's located in East Asia. 

\textit{Evidence:} Peking University abbreviated PKU is a major Chinese research university located in Beijing and a member of the C9 League.

the veracity of is deductively refuted. However, all the models labelled this pair as evidence supporting the claim. The rationale provided was that China was a sovereign country, ignoring the claim completely. This shows the over-reliance on specific parts of the evidence by the models while ignoring others.
Overall, our findings suggest that LLMs' reasoning capabilities are domain and task dependent.
Thus, we believe \benchmark represents a valuable resource to assess reasoning capabilities since it covers a wide spectrum of settings concerning claim verification.


\paragraph{Prompting Settings and Strategies}
Our experiments show that prompting strategies like ZS CoT and M-CoT do not lead to systematic performance improvements, but are rather specific to datasets (e.g., \vitc) and models.
These results align with recent findings about CoT being beneficial mainly for math- and code-related tasks~\citep{{sprague2024cotcotchainofthoughthelps}}.
This is likely derived by divergent reasoning paths within the models during inference that lead to reduction in performance~\citep{substackThinkAbout, Todd2023FunctionVI, Dutta2024HowTT}. Furthermore, we also observe that internal alignment can hinder reasoning capabilities when it comes to abductive reasoning. 
Models are averse to provide predictions when evidence is incomplete.
Yet abductive reasoning is often required for more complex tasks such as legal reasoning, just in time fact-checking, and other diverse forms of composite reasoning tasks. Hence, in order to achieve good results on these type of reasoning tasks, LLMs need to improve in the direction of abductive reasoning.


\paragraph{Explanation Quality} 
Our evaluation of generated explanations shows that these are on average consistent with human rationales.
In particular, ZS CoT rationales are more convincing due to their verbosity, whereas M-CoT rationales are more concise.
Moreover, we observe that rationales generated for abductive reasoning cases 
resemble assertions as models disprefer generating uncertain rationales.  
Nonetheless, considering that our results are limited to macro performance results and given the limited number of abductive cases, we believe our estimates to decrease as dataset size increases.
We leave a fine-grained analysis on generated rationales concerning an extended version of \benchmark as future work.


\section{Related work} \label{sec:related}

\paragraph{LLMs for Reasoning.}
Several contributions have evaluated different reasoning capabilities in LLMs, including atomic and compounds types.
For instance, LLMs can perform \textbf{abductive} reasoning for event prediction~\cite{shi2023language}, but struggle with common sense reasoning~\cite{zhao-etal-2023-abductive}.
Similarly, \textbf{deductive} reasoning in LLMs is beneficial to theorem-proving~\cite{saparov2023testing}, factual content generation~\cite{akyürek2024deductive}, and question-answering~\cite{li-etal-2024-look}. 
\citet{he2025towards} looked into analogical reasoning along with composite reasoning tasks and came to the conclusion that LLMs were able to handle structured analogical reasoning, but failed at more abstract composite reasoning tasks such as legal reasoning.
Nonetheless, the observed improvements are often attributable to how prompts are designed rather than an emergent deductive capability~\cite{chen-etal-2024-premise}.
Moreover, LLMs perform out-of-context \textbf{inductive} reasoning~\cite{treutlein-etal-2024-connecting}, but fail in lexical tasks~\cite{ye-etal-2023-assessing}.
Regarding \textbf{analogical} reasoning, LLMs address a wide variety of tasks, including nonverbal tests~\cite{Webb2023,hu-etal-2023-context}, question-answering~\cite{yu-etal-2023-pre}, mathematical problem solving~\cite{yasunaga-etal-2024-large}, and planning~\cite{yu-etal-2024-thought}, but present shortcomings in as many others~\cite{ye-etal-2024-analobench,sourati-etal-2024-arn,stevenson-etal-2024-large,ahrabian-etal-2024-curious,lewis-etal-2024-counterfactual}.
Likewise, despite promising results in \textbf{compound} reasoning tasks, such as counterfactual~\cite{wu2023reasoning}, 
and compositional reasoning~\cite{lu2023chameleon}, LLMs are notably unreliable~\cite{gao-etal-2023-chatgpt,zhang-etal-2024-probing}, sensitive to context~\cite{hosseini-etal-2024-not,chang-etal-2024-language}, and rely on shortcuts~\cite{yang-etal-2024-large-language-models}.

\paragraph{LLMs for Claim Verification.}
Early work with LLMs focused on verifying simple facts~\cite{lee2020language}.
More recently, LLMs for claim verification have been augmented with external knowledge~\cite{li2023selfchecker,factllama}, prompt-based reasoning~\cite{cao2023enhancing,li2023making,lin-etal-2023-beneath}, claim decomposition for fine-grained search into text chunks~\cite{li2023selfchecker} or first-order logic terms~\cite{wang-shu-2023-explainable}, and data-augmentation~\cite{alhindi2023large}.
A close work to ours is \cite{10.1109/TKDE.2025.3536008}.
Contrary to \citet{10.1109/TKDE.2025.3536008}, we evaluate models based on real-world claims and evidence.
Additionally, in \citet{10.1109/TKDE.2025.3536008}, models are prompted to generate outputs using specific reasoning modes. This likely implicitly biases the models and increases performance that might not measure proper understanding of the task.
By contrast, we evaluate the models by de-coupling these implicit signals and base our evaluation on understanding of the tasks.
Lastly, we also provide a framework for breaking down claim-evidence pairs into different atomic reasoning types.
While LLMs have been extensively applied in fact-checking, the question of which reasoning capabilities are needed to verify claims remains unaddressed.
Thus, we are the first to propose a reasoning benchmark for claim verification.
Although all three datasets used in our study can be used for fact checking, we do not explicitly cast our problem as improving fact checking performance.
For instance, \citet{hu-etal-2023-large} evaluates the models' fact checking capabilities based on models' internal knowledge using claims only. By contrast, we provide the model with claim and evidence pairs. Therefore, our focus centers more around the models' capabilities of understanding the evidence and implicitly reasoning to come to a conclusion in order to verify a claim.
Other examples are \cite{aly-etal-2023-qa}, \cite{tang-etal-2024-minicheck}, and \cite{strong-etal-2024-zero}, where a significant difference with our work is our choice of claims. 
In particular, all our claims are real-world claims and evidence.
Additionally, we focus on news articles as we build on the \pheme dataset rather than on Wikipedia sources. Furthermore, we also provide a reasoning framework to facilitate the decomposition of claims into atomic reasoning types.

\section{Conclusion}

We propose a novel extendable logical reasoning framework for deconstructing claim-evidence pairs into reasoning steps, required to determine the veracity of a claim.
We use our framework to create \benchmark, the first reasoning benchmark for claim verification focussed on deductive and abductive reasoning.
Our results show that LLMs notably struggle with abductive reasoning, while performing better in deductive cases.
Our findings show that LLMs reasoning capabilities are domain and task dependent.
In particular, no specific prompting strategy, including rationale generation, is systematically beneficial across all datasets and models.
Nevertheless, rationales generated by LLMs for deductive reasoning are on average consistent with human ones. 
Overall, these results highlight that \benchmark represents a challenging reasoning setting for LLMs and further research is required to reach satisfying performance.




\section*{Limitations}

\paragraph{Dataset Selection and Reasoning Types.}
Our focus on deductive and abductive reasoning types is dictated by our findings in the preliminary study.
Nonetheless, other resources could be investigated to expand our approach to include other reasoning types.
An example domain is biomedicine where datasets like COVID-Fact~\cite{saakyan-etal-2021-covid} could include examples where inductive reasoning is required to infer claim veracity.

\paragraph{Models.}
We analyse three widely adopted proprietary LLMs.
However, other models, including open-source ones, are also widely assessed in reasoning tasks.
For a broader evaluation of LLMs, our study could include other models although these are currently unlikely to outperform the most established proprietary models.

\paragraph{Rationale Generation.}
When LLMs generate an explanation, there is no guarantee that it is true to the final label assigned by the model. We mitigate this issue by obtaining both the label and explanation in the same prompt, although it should still be treated as merely "a plausible post-hoc explanation generated by the model" rather than the specific reason behind its decision.

\section*{Ethics Statement}

The \pheme dataset is a pre-existing dataset of rumours, for which ethical approval was obtained by the original research team. 
The rest of the datasets were sampled from pre-existing datasets for which no ethical approval was required. 

\section*{Acknowledgments}
This work was supported by a UKRI/EPSRC Turing AI Fellowship to Maria Liakata (grant ref EP/V030302/1) and the Alan Turing Institute (grant ref EP/N510129/1). This work was also supported by the Engineering and Physical Sciences Research Council [grant number EP/Y009800/1], through funding from Responsible Ai UK (KP0016) as a Keystone project lead by Maria Liakata. F. Ruggeri is partially supported by the project European Commission's NextGeneration EU programme, PNRR -- M4C2 -- Investimento 1.3, Partenariato Esteso, PE00000013 - ``FAIR - Future Artificial Intelligence Research'' -- Spoke 8 ``Pervasive AI’’ and by the European Union’s Justice Programme under Grant Agreement No. 101087342 for the project “Principles Of Law In National and European VAT”. Yulan He was supported by the UK Engineering and Physical Sciences Research Council through a Turing AI Fellowship (grant no. EP/V020579/1, EP/V020579/2).

\bibliography{bib}

\appendix
\label{sec:appendix}

\renewcommand{\thefigure}{A\arabic{figure}}
\setcounter{figure}{0}
\renewcommand{\thetable}{A\arabic{table}}
\setcounter{table}{0}

\section{Logical Reasoning Examples} \label{app:reasoning}

We hereby provide a formal description of atomic reasoning types, including examples on claim verification whenever possible.

\paragraph{Deductive.} 
\label{deducive_example}
Deductive reasoning or top-down logic is a logical reasoning process where we use inference rules such as modus ponens to deduce the veracity of a conclusion based on multiple hypotheses. 
%
%
A core element of deductive inference is that if the premises are true, then the conclusion is true. 
In formal logic, the rules of deduction are infinite \cite{pmlr-v202-morishita23a}, where the most common ones are modus ponens, syllogism, and elimination. 
The reader is referred to the works of \citet{pmlr-v202-morishita23a} and \citet{saparov2023testing} for a more in-depth discussion of deduction rules. \\ \\
\noindent \textbf{Example.}

\begin{itemize}[leftmargin=*,label={}]
   \item \underline{Claim:} \textit{Schools closed, Dammartin-en-Goele residents told to stay indoors, town `like warzone'.}
   \item \underline{Evidence:} \textit{Schools went into lockdown and the town appealed to residents to stay inside resident's houses.}
   \item \underline{Conclusion:} \textit{The evidence references the school closing down and residents being told to shelter at home. Therefore, we deductively infer that the rumour is true as the conclusion logically follows the evidence. }
\end{itemize}

\paragraph{Abductive.} 
There is much debate regarding definition of abductive reasoning \citep{Plutynski2011}. 
We follow the work of \citet{Paul1993}, which provides three different approaches towards defining abductive reasoning as:
\label{abductive_example}

\begin{itemize}
\item A set-cover-based approach; 
\item A logic-based approach;
\item A knowledge-level approach.
\end{itemize}

In this work, we use the set-cover-based approach, in which we construct the set of most plausible hypotheses $H$ given some observations $O$. 
Afterwards, we find the best possible explanation $E$ based on $H$. 
In other words, 

\begin{itemize}[leftmargin=*,label={}]
   \item `\textit{A domain for hypothesis assembly is defined by the triple $\phi$, $\sigma$, $\epsilon$), where $\phi$ is a finite set of hypotheses, $\sigma$ is a set of observations and $\epsilon$ is a mapping from subsets of $\phi$ to subsets of $\sigma$. $\epsilon(\phi)$ is called the explanatory power of the set of hypotheses $\phi$ and determines the set of observations $\sigma$ accounts for. An assembly problem is given by a set $\sigma' \subseteq \sigma$ of observations that have to be explained.}' \citep{Paul1993}. 
\end{itemize}

Additionally, the key difference between abductive reasoning and the other forms of reasoning types is that, unlike the other types, abdctive reasoning works "backwards" towards the most plausible hypothesis from a given set of rules and happenings. Deductive reasoning is formulation of results based on rule and observation and inductive reasoning is formulation of rule based on result and observation. Whereas, abductive reasoning is formulation of an observation based on rule and result. 
For example from \citet{Flach2000}:

\begin{description}
    \item \underline{Rule} All the beans from this bag are white. 
    \item \underline{Result} These beans are white. 
    \item \underline{Conclusion} These beans are from this bag.
\end{description}

\paragraph{Inductive.} 
Inductive reasoning is the reasoning process where we use observations and outcomes to infer a generalizable rule. Hence, the logical structure can be represented as:

\begin{itemize}[leftmargin=*,label={}]
\item $\forall x, observations(x) \implies conclusion$ 

or

\item $\exists x, observations(x) \implies conclusion$ 
\end{itemize}

amongst many other forms. 
A conclusion reached by inductive reasoning is not necessarily true. 
As per \citet{Flach2000}, if the premises for any stated argument only provide partial support for its conclusion, then that argument is inductive supposing the premises are true. \\

\noindent \textbf{Example 1.}

\begin{itemize}[leftmargin=*,label={}]
   \item \underline{Claim:} \textit{Injecting or consuming bleach is good for killing the virus (Covid-19).} 
   \item \underline{Evidence 1:} \textit{Applying bleach or chlorine to the skin can cause harm, especially if it enters the eyes or mouth. }
   \item \underline{Evidence 2:} \textit{These chemicals can disinfect surfaces, but people should not use them on their bodies.}
   \item \underline{Evidence 3:} \textit{Also, these products cannot kill viruses inside the body.}
   \item \underline{Conclusion:} \textit{From the evidence we can inductively draw a general conclusion that the claim is false, as bleach causes harm to the body and would not kill any viruses within.}
\end{itemize}

\noindent \textbf{Example 2.}

\begin{description}
    \item \underline{Observation1} Eagles have wings. Eagles are birds and eagles can fly. 
    \item \underline{Observation2} Ducks have wings. Ducks are birds and ducks can fly. 
    \item \underline{Observation 3a} Pigeons have wings. Pigeons are birds and pigeons can fly.
    
    or
    
    \item \underline{Observation 3b} Bats have wings. Bats are mammals and bats can fly.
    \item \underline{Conclusion a} All birds have wings and all birds can fly.
    
    or
    
    \item \underline{Conclusion b} Those who have wings can fly.
\end{description}

It is clear that each conclusion is true if we make a closed-world assumption regarding the premises. However, in reality, it is false as there exist flightless birds including Penguins and wingless birds such as Kiwi.

\paragraph{Analogical.} 
Analogical reasoning is the reasoning process concerned with comparison between two or more objects, arguments, or entities. \\

\noindent \textbf{Example.}

\begin{itemize}[leftmargin=*,label={}]
    \item \underline{Claim:} entity $\alpha$ is equivalent to entities $\zeta$, $\kappa$, $\phi$, and $\omega$.

    \item \underline{Evidence:} entity $\beta$ is equivalent to entities $\zeta$, $\kappa$, and $\phi$.

    \item \underline{Conclusion:} entity $\beta$ is probably equivalent to entity $\omega$.
\end{itemize}

\section{Claim Verification Datasets} \label{app:datasets}

We select three popular resources for claim verification, covering different domains and increasing task complexity.


\paragraph{\vitc~\cite{schuster-etal-2021-get}.}
A multi-task fact-checking dataset based on manual and synthetic English revisions to  Wikipedia pages.
The dataset comprises $\sim$450k claim-evidence pairs.
For the claim verification task, claim-evidence pairs are annotated with veracity labels: \textit{supports}, \textit{refutes}, and \textit{not-enough-information}. VitaminC is licensed under MIT License.


\paragraph{\climate~\cite{diggelmann2020climatefever} .}
A claim verification dataset that consists of $\sim$1.5k real-world claims concerning climate change.
The claims are retrieved from Google while the evidence is Wikipedia-based.
The claim-evidence pairs are annotated with veracity labels: \textit{supports}, \textit{refutes}, and \textit{not-enough-information}. 

\paragraph{\pheme~\cite{phemeplus}.}
A rumour verification dataset comprising social media claims about real-world events. 
The dataset contains five different events where associated claim-evidence pairs are annotated with veracity labels: \textit{true}, \textit{false}, \textit{not-enough-information}.
\pheme is an extension of the PHEME~\cite{pheme} dataset, where web-retrieved news articles are used as evidence in place of Twitter threads as done in PHEME.


\section{Sampling details}
\label{sec:appx_sample}

\begin{figure*}[!hb]
\centering
\includegraphics[width=1.2\linewidth]{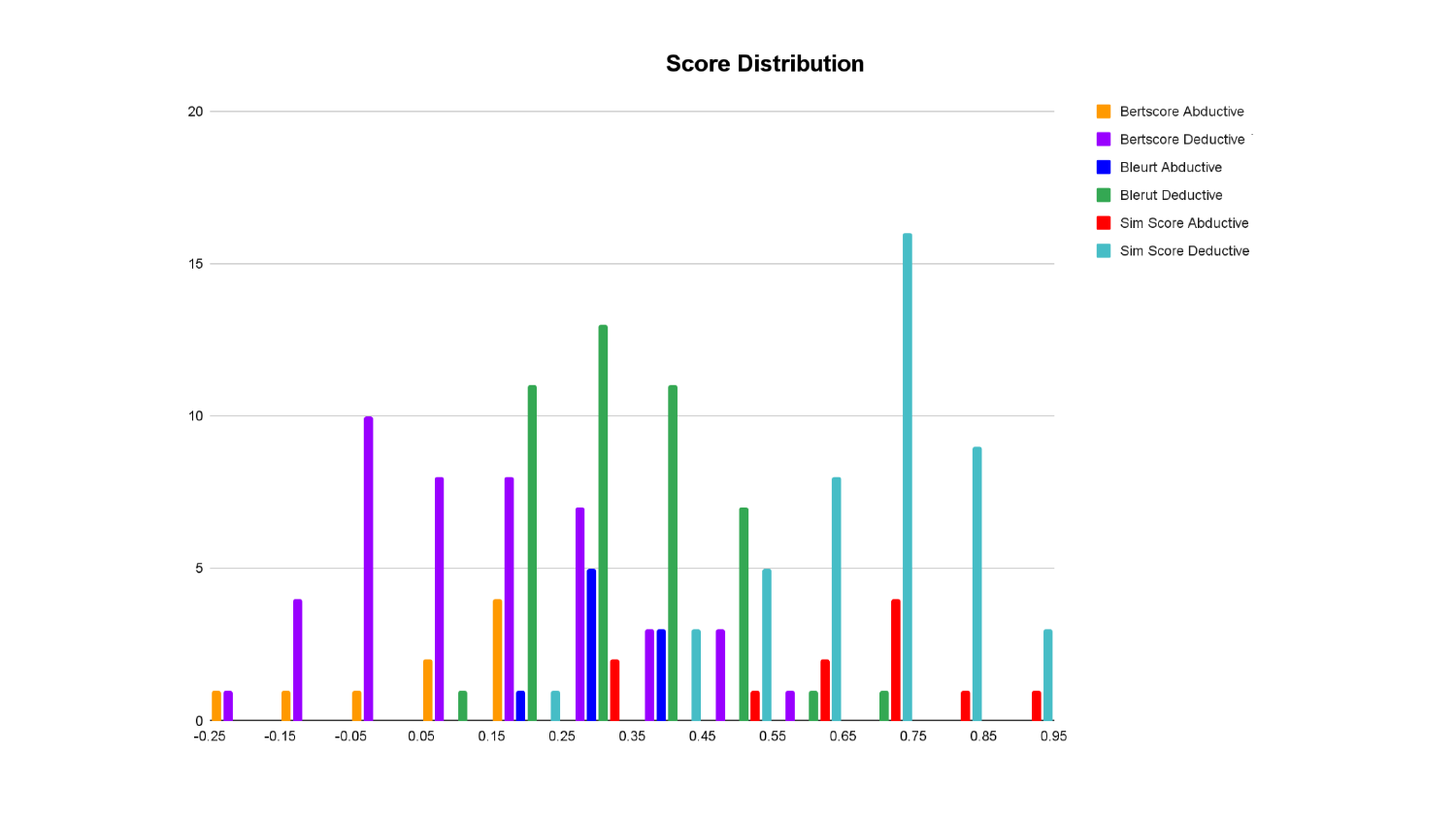}
\caption{\label{fig:sampline_threshold} Distribution of the sampling metrics.}
\end{figure*}

Figure~\ref{fig:sampline_threshold} shows the distribution of cosine similarity (denoted as Sim Score), BERTScore and BLEURT score. We used this distribution to set up two different thresholds for deductive and abductive samples. 
We found that the Bertscores and Sim Scores for abductive sample did not seem to overlap. However, the deductive score had overlap between them. From this observation, we derived the following thresholds.


\section{Annotation Guidelines} \label{app:guidelines}

Figure~\ref{fig:methodology} summarizes our annotation pipeline for \benchmark. In data annotation (Figure~\ref{fig:methodology}, \textbf{Top}), we provide a human annotator with claim-evidence pairs with corresponding veracity label.
The annotator determines the reasoning type required to infer the claim veracity and provides a rationale in free-text format as motivation.
\input{figures/methodology}
Table~\ref{tab:guidelines} reports the annotation guidelines we used to instruct annotators in creating \benchmark.

\input{tables/guidelines}


\section{Data Annotation} \label{app:data-annotation}

Table~\ref{tab:pairwise-agreement} reports pairwise agreement scores for each dataset in \benchmark.

\input{tables/agreement}

\section{Prompts} \label{app:prompts}

Tables~\ref{tab:vitc-prompts}, \ref{tab:climate-prompts}, and \ref{tab:pheme-prompts} report the prompts we used for \vitc, \climate, and \pheme, respectively.
We follow standard prompt construction strategies and provide dataset specific personas and instructions.
Additionally, in M-CoT with provide examples to guide the model.



\input{tables/vitc_prompts}
\input{tables/climate_prompts}
\input{tables/pheme_prompts}

\section{Human Rationale Evaluation} \label{app:human-rationale}

A third annotator judges the quality of the rationales based on coherence, relevance, and consistency.
We denote this annotator as \textit{evaluator} for readability.
The evaluator only considers rationales where annotators agree with each other. 
In particular, the evaluator selects rationales on the basis of containing the most relevant and coherent information, and providing a consistent narrative.
For samples where annotators disagree, the rationale of the annotator with most rationales within the agreement set was considered.

\section{Qualitative Analysis} \label{app:qualitative analysis}
Tables~\ref{tab:pairwise_permutation_vitc}, \ref{tab:pairwise_permutation_climate}, and \ref{tab:pairwise_permutation_pheme} report pairwise permutation tests on \benchmark datasets.
Moreover, Tables~\ref{tab:vitc_evaluation}, \ref{tab:climate_evaluation}, and \ref{tab:pheme_evaluation} report qualitative analysis metrics on \benchmark datasets.
In particular, we compute qualitative metrics on two sets of examples: those for which models correctly predicted the corresponding claim veracity (\textbf{Correct}), and those where models made wrong predictions (\textbf{Wrong}).

The metrics used for qualitative analysis are as following, 

\paragraph{Factual consistency.} 
We assess the consistency of LLM generated rationales $R$ with human-written ones $H$, where consistency is the absence of contradiction. We define $C$ to be a function that quantifies the consistency of text $B$ based on text $A$: 

\begin{equation*} 
\resizebox{1.05\hsize}{!}{$
C(A, B) = \frac{1}{|A| \cdot |B|} \sum_{a \in A} \sum_{b \in B} \left(1 - \text{NLI}(\text{Contradict}|a, b)\right)
$}
\end{equation*}

\noindent We calculate the consistency of LLM rationales to human rationales as, 

\begin{equation}
\text{FC} = 1 - \frac{1}{N} \sum_{i=1}^{N} C_i
\end{equation}

where $N$ is the total number of sentence pairs compared,$C_i$ is the consistency score of the $i$-th comparison.

\paragraph{Evidence appropriateness.} 

For evidence appropriateness, we use the same consistency score $C$ as Fact\_Expert. 
\begin{equation}
\text{EA} = \frac{1}{M} \sum_{j=1}^{M} \left( \frac{1}{N_j} \sum_{i=1}^{N_j} (1 - c_{ij}) \right)
\end{equation}

Here, $M$ is the total number of generated rationales, $N_j$ is the number of sentences in the $j$-th generated rationale and $C_ij$ is the consistency score for the $i$-th sentence in the $j$-th rationale. Evidence appropriateness can be considered as the mean factual consistency whereas Fact\_Expert is the granular sentence level consistency. 


\paragraph{Coherence.} 
We estimate how easy it is to follow the rationales and how effectively it integrates information from the evidence using BARTScore.


\paragraph{Fluency.}  
We estimate fluency for rationales using perplexity (PPL) under \textsc{GPT-2-xl} \cite{radford2019language}.

\input{tables/evaluation_vitc}
\input{tables/evaluation_climate}
\input{tables/evaluation_pheme}

\input{tables/permutation_test_vitc}
\input{tables/permutation_test_climate}
\input{tables/permutation_test_pheme}

\section{Statistical Significance Tests} \label{app:statistical-significance}

We run a non-parametric Mann Whitney U test \cite{mcknight2010mann} as it has no normality assumption and works with unequal population sizes.
Table \ref{tab:mwu} reports results.
Our hypotheses are as follows.

\paragraph{Null Hypothesis ($H_0$).}
There's no difference between the distributions in accuracy when predicting veracity for deductive and abductive reasoning.

\paragraph{Alternative Hypothesis ($H_1$).}
There is a difference — the model behaves differently on abductive reasoning cases.

\begin{table}[!tb]
    \small
    \centering
    \resizebox{\columnwidth}{!}{%
    \begin{tabular}{l|ccc}
    \toprule
     \textbf{Model}  &  \vitc & \climate & \pheme \\ \midrule
      Claude ZS   & $7.19e^{-4}$ & $1.35e^{-12}$ & $2.61e^{-3}$ \\
      Claude ZS CoT & $2.42e^{-3}$ & $7.32e^{-11}$ & $5.35e^{-3}$ \\
      Claude M-CoT & $2.45e^{-4}$ & $1.00e^{-17}$ & $1.51e^{-2}$ \\
      GPT-4 ZS & $4.44e^{-5}$ & $1.78e^{-11}$ & $1.31e^{-5}$ \\
      GPT-4 ZS CoT & $2.08e^{-5}$ & $4.20e^{-12}$ & $1.24e^{-4}$ \\
      GPT-4 M-CoT & $3.25e^{-6}$ & $1.39e^{-8}$ & $4.14e^{-5}$ \\
      GPT-4o ZS & $3.75e^{-4}$ & $7.56e^{-10}$ & $2.45e^{-4}$ \\
      GPT-4o ZS CoT & $4.63e^{-4}$ & $7.32e^{-11}$ & $3.75e^{-4}$ \\
      GPT-4o M-CoT & $4.35e^{-6}$ & $5.08e^{-10}$ & $8.65e^{-5}$ \\ \bottomrule
    \end{tabular}%
    }
    \caption{Two-sided Mann Whitney U test results.}
    \label{tab:mwu}
\end{table}

\end{document}

%% file: figures/framework.tex
\begin{figure*}[!htb]
\centering
\includegraphics[width=0.7\linewidth]{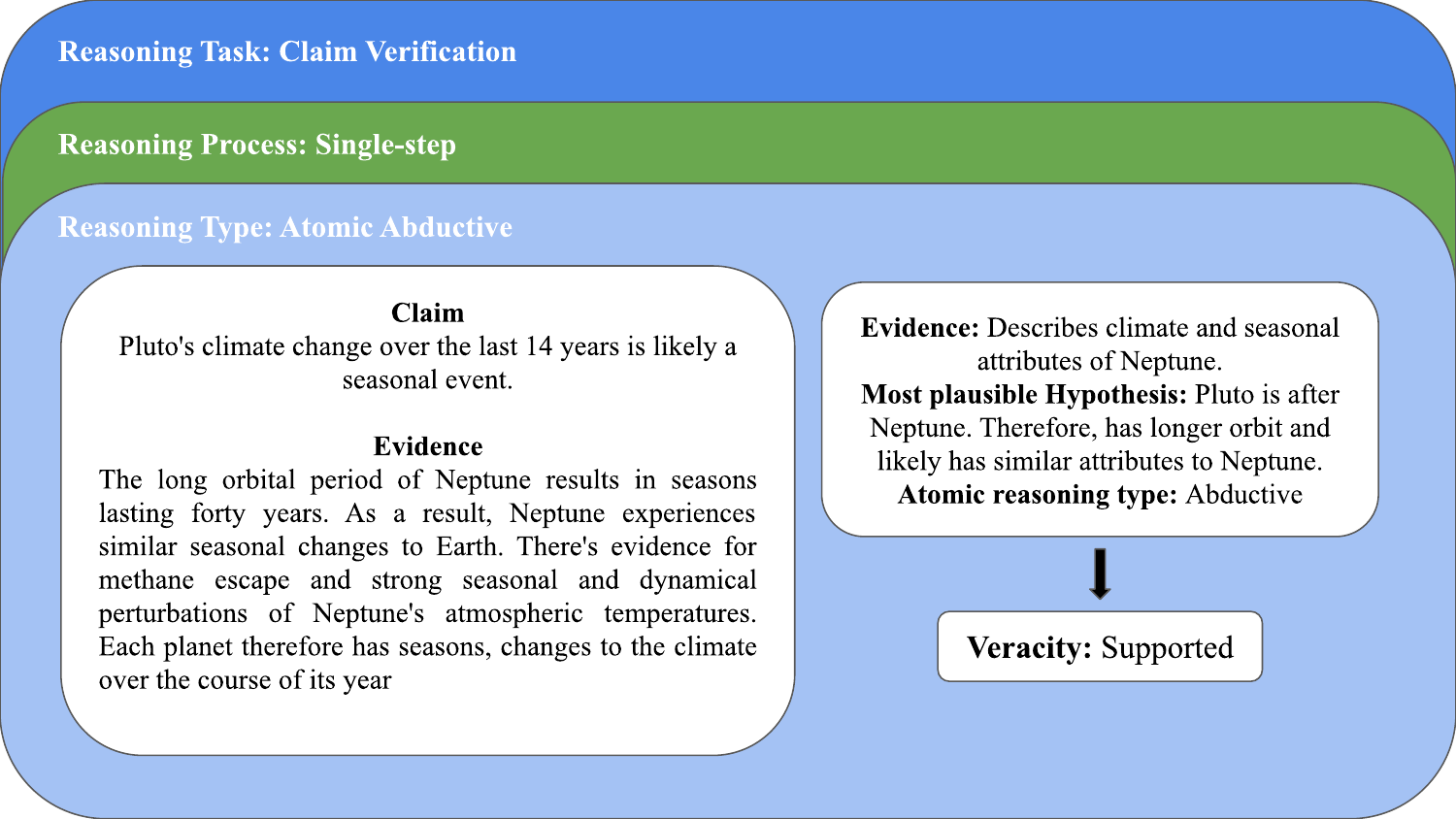}
\caption{\label{fig:reasoning_path} Resolution of claim verification via a single-step abductive reasoning type using \benchmark framework.}
\end{figure*} 

%% file: tables/benchmark_stats.tex
\begin{table}[!t]
\small
\begin{center}
\begin{adjustbox}{max width=\textwidth}
\begin{tabular}{@{}lccc@{}}
\toprule
\textbf{\vitc}       & \textbf{Supported} & \textbf{Refuted} & \textbf{Total} \\
\midrule
Deductive               & 272                & 199              & 471            \\
Abductive               & 11                 & 18               & 29             \\
\textbf{Total}          & 283                & 217              & 500   \\ \midrule
\textbf{\climate} & \textbf{Supported} & \textbf{Refuted} & \textbf{Total} \\
\midrule
Deductive               & 269                & 129              & 398            \\
Abductive               & 88                 & 14               & 102            \\
\textbf{Total}          & 357                & 143              & 500   \\
\midrule
\textbf{\pheme} & \textbf{Supported} & \textbf{Refuted} & \textbf{Total} \\
\midrule
Deductive               & 336                & 128              & 464            \\
Abductive               & 22                 & 14               & 36             \\
\textbf{Total}          & 358                & 142              & 500   \\
\bottomrule
\end{tabular}
\end{adjustbox}
\end{center}
\caption{\benchmark statistics.}
\label{tab:claim_distribution}
\end{table}

%% file: tables/merged_results.tex
\begin{table*}[!t]
\begin{center}
\begin{adjustbox}{max width=\textwidth}
\begin{tabular}{@{}l rrr rrr rrr@{}}
\toprule
 & \multicolumn{3}{c}{\textbf{\vitc}} & \multicolumn{3}{c}{\textbf{\climate}} & \multicolumn{3}{c}{\textbf{\pheme}} \\
\cmidrule(r){2-4} \cmidrule(r){5-7} \cmidrule(r){8-10}
\multicolumn{1}{l}{\textbf{Model}} & \textbf{F1} $\uparrow$ & \textbf{Deductive} $\downarrow$ & \textbf{Abductive} $\downarrow$ & \textbf{F1} $\uparrow$ & \textbf{Deductive} $\downarrow$ & \textbf{Abductive} $\downarrow$ & \textbf{F1} $\uparrow$ & \textbf{Deductive} $\downarrow$ & \textbf{Abductive} $\downarrow$ \\
\midrule
Claude ZS \emph{No-Exp} & 0.85 & 13.62\alignempty & 33.33\alignempty & 0.80 & 12.81\alignempty & 40.20\alignempty & 0.73 & 19.40\alignempty & \textbf{38.89}\alignempty \\
Claude M-CoT \emph{No-Exp} & 0.87 & 12.77\alignempty & \textbf{23.33}\alignempty & 0.80 & 12.81\alignempty & 41.84\alignempty & \textbf{0.76} & \underline{18.53}\alignempty & \textbf{38.89}\alignempty \\

GPT-4 ZS \emph{No-Exp} & 0.86 & 12.13\alignempty & 30.00\alignempty & \underline{0.87} & \textbf{8.79}\alignempty & \textbf{20.59}\alignempty & 0.69 & 20.69\alignempty & \textbf{38.89}\alignempty \\
GPT-4 M-CoT \emph{No-Exp} & \textbf{0.90} & \textbf{8.30}\alignempty & \underline{26.67}\alignempty & 0.85 & 10.05\alignempty & 27.45\alignempty & 0.70 & 22.41\alignempty & 52.78\alignempty \\

GPT-4o ZS \emph{No-Exp} & 0.88 & 10.43\alignempty & 40.00\alignempty & 0.84 & 9.55\alignempty & 33.33\alignempty & 0.72 & 20.04\alignempty & \underline{41.67}\alignempty \\
GPT-4o M-CoT \emph{No-Exp} & 0.88 & 10.43\alignempty & 30.00\alignempty & \textbf{0.92} & \underline{9.05}\alignempty & \underline{25.49}\alignempty & 0.74 & 19.40\alignempty & 47.22\alignempty \\
\midrule
Claude ZS \emph{Exp} & 0.89 & 9.79$^{{(\posdelta{+3.83})}}$ & 30.00$^{{(\posdelta{+3.33})}}$ & 0.74 & 17.34$^{(\negdelta{-4.53})}$ & 52.94$^{(\negdelta{-12.75})}$ & 0.74 & 20.04$^{(\negdelta{-0.65})}$ & \underline{41.67}$^{(\negdelta{-2.78})}$ \\
Claude ZS CoT \emph{Exp} & 0.88 & 11.06\alignempty & 30.00\alignempty & 0.70 & 22.61\alignempty & 56.86\alignempty & 0.73 & 21.34\alignempty & 41.67\alignempty \\
Claude M-CoT \emph{Exp} & \underline{0.90} & \underline{8.72}$^{{(\posdelta{+4.04})}}$ & 30.00$^{{(\negdelta{-6.67})}}$ & 0.73 & 17.59$^{(\negdelta{-4.77})}$ & 57.84$^{(\negdelta{-16.01})}$ & 0.73 & 23.49$^{(\negdelta{-4.96})}$ & \underline{41.67}$^{(\negdelta{-2.78})}$ \\

GPT-4 ZS \emph{Exp} & 0.88 & 10.64$^{{(\posdelta{+1.49})}}$ & 36.67$^{{(\negdelta{-6.67})}}$ & 0.78 & 15.08$^{(\negdelta{-6.28})}$ & 47.06$^{(\negdelta{-26.47})}$ & 0.73 & 20.04$^{(\posdelta{+0.65})}$ & 52.78$^{(\negdelta{-13.89})}$ \\
GPT-4 ZS CoT \emph{Exp} & 0.88 & 10.00\alignempty & 36.67\alignempty & 0.77 & 14.32\alignempty & 57.84\alignempty & 0.71 & 21.12\alignempty & 50.00\alignempty \\
GPT-4 M-CoT \emph{Exp} & 0.89 & 8.72$^{{(\negdelta{-0.42})}}$ & 36.67$^{{(\negdelta{-10.00})}}$ & 0.82 & 11.31$^{(\negdelta{-1.26})}$ & 35.29$^{(\negdelta{-7.84})}$ & 0.73 & 19.61$^{(\posdelta{+2.80})}$ & 50.00$^{(\posdelta{+2.78})}$ \\

GPT-4o ZS \emph{Exp} & 0.89 & 9.15$^{{(\posdelta{+1.28})}}$ & 30.00$^{{(\posdelta{+10.00})}}$ & 0.79 & 14.07$^{(\negdelta{-4.52})}$ & 42.16$^{(\negdelta{-8.82})}$ & 0.74 & \textbf{18.32}$^{(\posdelta{+1.72})}$ & 44.44$^{(\negdelta{-2.78})}$ \\
GPT-4o ZS CoT \emph{Exp} & 0.89 & 9.36\alignempty & 30.00\alignempty & 0.78 & 14.07\alignempty & 55.88\alignempty & 0.74 & 18.97\alignempty & 44.44\alignempty \\
GPT-4o M-CoT \emph{Exp} & 0.89 & 8.94$^{{(\posdelta{+9.96})}}$ & 36.67$^{{(\negdelta{-6.67})}}$ & 0.78 & 13.82$^{(\negdelta{-4.77})}$ & 42.16$^{(\negdelta{-16.67})}$ & \underline{0.75} & 18.75$^{(\posdelta{+0.65})}$ & 47.22$^{(+0.00)}$ \\

\bottomrule
\end{tabular}
\end{adjustbox}
\end{center}
\caption{Claim verification performance on \benchmark. Best results are in \textbf{bold}, second-best results are \underline{underlined}. We report error rate delta performance between \emph{No-Exp} and \emph{Exp} settings in brackets. Negative delta indicates that rationale generation degrades perforamnce.}
\label{tab:claim-verification}
\end{table*}

%% file: tables/human_evaluation.tex
\begin{table*}[!t]
\begin{center}
\begin{adjustbox}{max width=\textwidth}
\small
\begin{tabular}{@{}l rrrr rrrr rrrr@{}}
\toprule
 & \multicolumn{4}{c}{\textbf{\vitc}} & \multicolumn{4}{c}{\textbf{\climate}} & \multicolumn{4}{c}{\textbf{\pheme}} \\
\cmidrule(r){2-5} \cmidrule(r){6-9} \cmidrule(r){10-13}
\multicolumn{1}{l}{\textbf{Model}} & \textbf{EA} $\uparrow$ & \textbf{FC} $\uparrow$ & \textbf{BART} $\uparrow$ & \textbf{PPL} $\downarrow$ & \textbf{EA} $\uparrow$ & \textbf{FC} $\uparrow$ & \textbf{BART} $\uparrow$ & \textbf{PPL} $\downarrow$ & \textbf{EA} $\uparrow$ & \textbf{FC} $\uparrow$ & \textbf{BART} $\uparrow$ & \textbf{PPL} $\downarrow$ \\
\midrule
Claude ZS & 0.85 & 0.85 & -4.16 & 99.63 & 0.87 & 0.88 & -4.26 & 31.08 & 0.82 & 0.81 & -4.31 & 39.82 \\
Claude ZS CoT & 0.82 & 0.83 & -4.38 & 52.85 & 0.82 & 0.83 & -4.28 & 29.57 & 0.85 & 0.84 & -4.44 & 38.81 \\
Claude M-CoT & 0.85 & 0.86 & -4.05 & 68.53 & 0.89 & 0.90 & -3.42 & \underline{25.52} & \textbf{0.89} & \underline{0.88} & -4.17 & \underline{37.83} \\

GPT-4 ZS & 0.87 & 0.87 & -3.83 & 66.84 & 0.91 & 0.91 & -3.67 & 27.93 & \underline{0.87} & 0.86 & -3.89 & 40.83 \\
GPT-4 ZS CoT & 0.87 & 0.86 & -3.78 & 59.01 & 0.90 & 0.91 & -3.65 & \textbf{20.65} & \underline{0.87} & 0.87 & -3.89 & \textbf{28.83} \\
GPT-4 M-CoT & 0.89 & \underline{0.88} & \textbf{-2.98} & \underline{45.84} & \textbf{0.93} & \textbf{0.94} & \textbf{-2.90} & 28.13 & 0.85 & 0.85 & \textbf{-3.40} & 47.15 \\

GPT-4o ZS & \underline{0.90} & \underline{0.88} & -3.63 & 52.96 & \underline{0.92} & \underline{0.93} & -3.64 & 58.42 & 0.85 & 0.86 & -4.01 & 99.63 \\
GPT-4o ZS CoT & \textbf{0.91} & \textbf{0.89} & \underline{-3.45} & \textbf{35.82} & \textbf{0.93} & \textbf{0.94} & \underline{-3.39} & 46.92 & \textbf{0.89} & \textbf{0.90} & \underline{-3.74} & 52.85 \\
GPT-4o M-CoT & \underline{0.90} & \underline{0.88} & -3.63 & 50.10 & 0.90 & 0.91 & -3.57 & 57.65 & \underline{0.87} & 0.86 & -4.08 & 68.53 \\

\bottomrule
\end{tabular}
\end{adjustbox}
\end{center}
\caption{Qualitative evaluation on \benchmark in the \emph{Exp} setting. Best results are in \textbf{bold}, second-best results are \underline{underlined}.}
\label{tab:human-evaluation}
\end{table*}

%% file: figures/methodology.tex
\begin{figure*}[!t]
\centering
\includegraphics[width=0.8\linewidth]{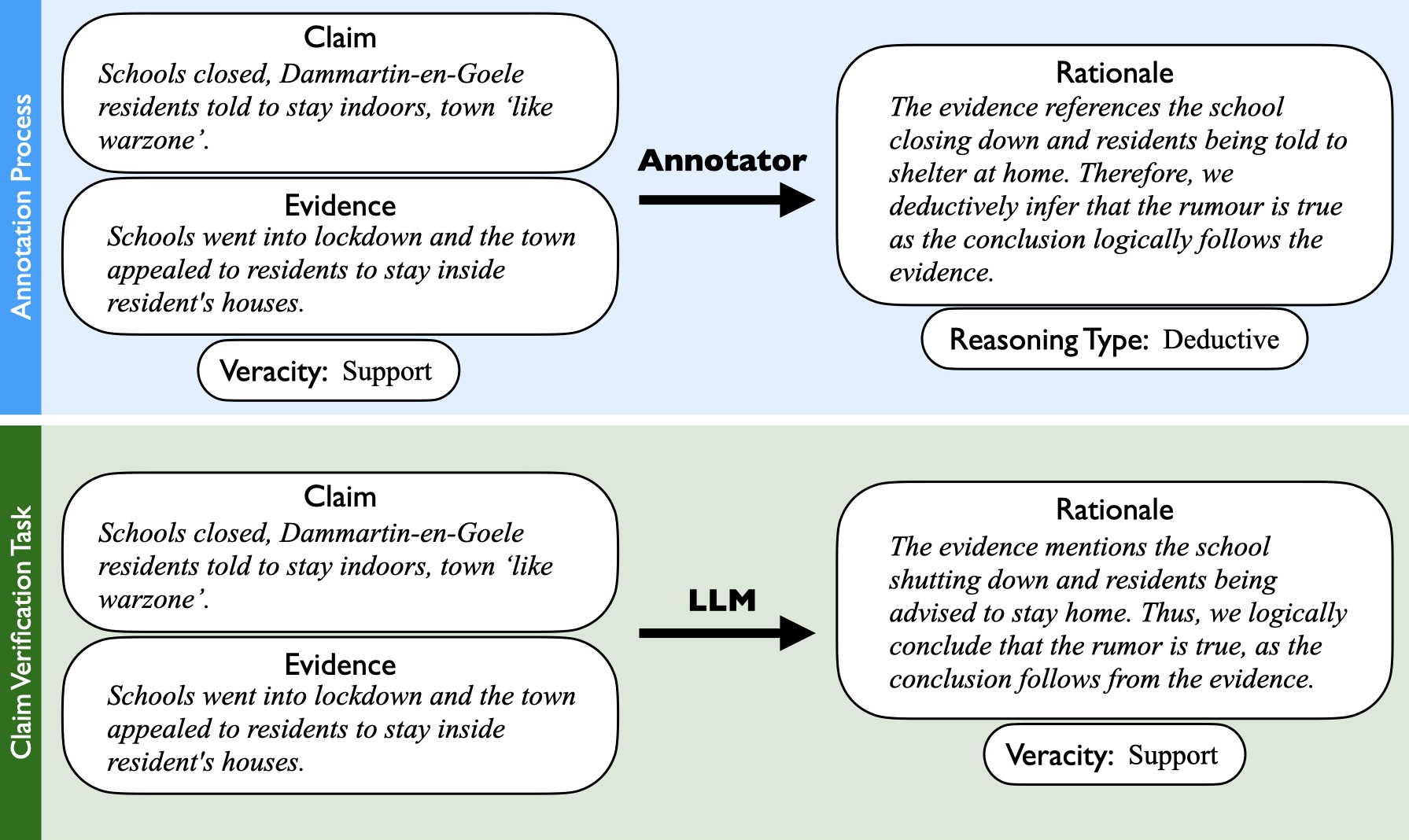}
\caption{\label{fig:methodology} (\textbf{Top}) Our annotation process for reasoning-based claim verification. An annotator provides reasoning type required to infer the claim veracity and a rationale as motivation. (\textbf{Bottom}) The claim verification task where a LLM has to predict the claim veracity and generate a rationale as support.}
\end{figure*} 

%% file: tables/guidelines.tex
\begin{table*}[!t]
\centering
\begin{tabular}{|p{2.0\columnwidth}|}
\hline 
Read each claim and evidence pair samples along with their associated veracity labels. Afterwards you will label them with the type of reasoning you think was necessary for inferring the veracity label of the claim given the evidence. The reasoning types are abductive and deductive. Also, provide rationale for your labels.
\\ \\
The goal is to identify what type of reasoning is necessary to infer the veracity of the claim given the associated evidence, for each of the given pairs. 
\\ \\
\textbf{Example 1.} \\ \\

\textbf{Claim:} Climate change isn't increasing extreme weather damage costs. \\ \\

\textbf{Evidence:} 1. Many analyses, such as that of the Stern Review presented to the British Government, have predicted reductions by several percent of world gross domestic product due to climate related costs such as dealing with increased extreme weather events and stresses to low-lying areas due to sea level rises. 2. Global losses reveal rapidly rising costs due to extreme weather-related events since the 1970s. 3. Global warming boosts the probability of extreme weather events, like heat waves, far more than it boosts more moderate events. 4. "Impacts [of climate change] will very likely increase due to increased frequencies and intensities of some extreme weather events". \\ \\

\textbf{Veracity:} Refutes \\ \\

\textbf{Reasoning:} Deductive \\ \\

\textbf{Rationale:} The evidence deductively refutes the claim. We find explicit mention of increased damage cost in the second line of the evidence. While the last two lines of evidence provide explicit evidence of global causing more adverse weather events. \\ \\

Example 2. \\ \\

\textbf{Claim}: Pluto's climate change over the last 14 years is likely a seasonal event. \\ \\

\textbf{Evidence}: The long orbital period of Neptune results in seasons lasting forty years. 2. As a result, Neptune experiences similar seasonal changes to Earth. 3. "Evidence for methane escape and strong seasonal and dynamical perturbations of Neptune's atmospheric temperatures". 4. Each planet therefore has seasons, changes to the climate over the course of its year. \\ \\

\textbf{Veracity}: Supports \\ \\

\textbf{Reasoning}: Abductive \\ \\

\textbf{Rationale}: The claim is abductively supported. Given Pluto used to be a planet and now is labeled as a dwarf planet, we can hypothesize that it likely has the same attribute as neptune. Given pluto has the biggest orbital period, it is very much likely pluto seasons last over 10 years. \\

\hline

\end{tabular}
\caption{Annotation guidelines used to create \benchmark. This specific example was for CLIMATE-FEVER. This partial representation of the guideline as we provided 8 examples for each dataset with a mix of supports and refutes label.}
\label{tab:guidelines}
\end{table*}

%% file: tables/agreement.tex
\begin{table*}[!t]
\begin{center}
\begin{adjustbox}{max width=\textwidth}
\begin{tabular}{@{}l|ccc|ccc|ccc@{}}
\toprule
 & \multicolumn{3}{c|}{\textbf{\vitc}} & \multicolumn{3}{c|}{\textbf{\climate}} & \multicolumn{3}{c}{\textbf{\pheme}} \\
\cmidrule(r){2-4} \cmidrule(r){5-7} \cmidrule(r){8-10}
 & \textbf{Annotator A} & \textbf{Annotator B} & \textbf{Annotator C} & \textbf{Annotator A} & \textbf{Annotator B} & \textbf{Annotator C} & \textbf{Annotator A} & \textbf{Annotator B} & \textbf{Annotator C} \\
\midrule
\textbf{Annotator A} & - & 0.72 & 0.74 & - & 0.56 & 0.56 & - & 0.64 & 0.68 \\
\textbf{Annotator B} & 0.72 & - & 0.78 & 0.56 & - & 0.56 & 0.64 & - & 0.68 \\
\bottomrule
\end{tabular}
\end{adjustbox}
\end{center}
\caption{Pairwise Bennett's S Score across different datasets.}
\label{tab:pairwise-agreement}
\end{table*}

%% file: tables/vitc_prompts.tex
\begin{table*}[!t]
\centering
\begin{tabular}{|p{2.0\columnwidth}|}
\hline

\textbf{ZS.} \\
You are an expert fact checker. 
As an expert fact checker, you will be helping us verify some claims. 
For your task, you will be provided with claims and evidence in this format Q:[<!C> Claim: ...<C!> \textbackslash n <!E> Evidence: ...<E!>]. 
You will use the provided evidence to decide whether the associated claim is supported or refuted. 
You will first briefly explain your reasoning in one sentence, and then make the final judgment by writing LABEL: followed by a single word SUPPORTS or REFUTES. \\ \\

\textbf{CoT.} \\
You are an expert fact checker. 
As an expert fact checker, you will be helping us verify some claims. 
For your task, you will be provided with claims and evidence in this format Q:[<!C> Claim: ...<C!> \textbackslash n <!E> Evidence: ...<E!>]. 
You will use the provided evidence to decide whether the associated claim is supported or refuted. 
You will first briefly explain your reasoning in one sentence, and then make the final judgment by writing LABEL: followed by a single word SUPPORTS or REFUTES.
Let's think step by step. \\ \\

\textbf{M-CoT.} \\
You are an expert fact checker. 
As an expert fact checker, you will be helping us verify some claims. 
You will be provided with tuples of claim, evidence and answer as examples first. 
The example claims will be inside <!eC>...<eC!> tokens, evidence will be inside <!eE>...<eE!> tokens and the answer/reasoning will be inside <!eA>...<eA!> tokens.
The answer is based on the evidence and it verifies whether the evidence supports or refutes the claim. 
For your task, you will be provided with claims and evidence in this format Q:[<!C> Claim: ...<C!> \textbackslash n <!E> Evidence: ...<E!>]. 
You will use the provided evidence to decide whether the associated claim is supported or refuted. 
You will first briefly explain your reasoning in one sentence, and then make the final judgement by writing LABEL: followed by a single word SUPPORTS or REFUTES. \\ \\

Here are some examples: \\
\{\texttt{examples}\} \\ \\

\hline

\end{tabular}
\caption{Prompts used in \vitc.}
\label{tab:vitc-prompts}
\end{table*}

%% file: tables/climate_prompts.tex
\begin{table*}[!t]
\centering
\begin{tabular}{|p{2.0\columnwidth}|}
\hline

\textbf{ZS.} \\
You are an expert climate scientist. 
As an expert climate scientist, you will be helping us verify some climate-related claims. 
For your task, you will be provided with climate-related claims and evidence in this format Q:[<!C> Claim: ...<C!> \textbackslash n <!E> Evidence: ...<E!>]. 
You will use the provided evidence to decide whether the associated claim is supported or refuted. 
You will first briefly explain your reasoning in one sentence, and then make the final judgement by writing LABEL: followed by a single word SUPPORTS or REFUTES. \\ \\

\textbf{CoT.} \\
You are an expert climate scientist. 
As an expert climate scientist, you will be helping us verify some climate-related claims. 
For your task, you will be provided with climate-related claims and evidence in this format Q:[<!C> Claim: ...<C!> \textbackslash n <!E> Evidence: ...<E!>]. 
You will use the provided evidence to decide whether the associated claim is supported or refuted. 
You will first briefly explain your reasoning in one sentence, and then make the final judgement by writing LABEL: followed by a single word SUPPORTS or REFUTES. 
Let's think step by step. \\ \\

\textbf{M-CoT.} \\
You are an expert climate scientist. 
As an expert climate scientist, you will be helping us verify some climate-related claims. 
You will be provided with tuples of claim, evidence and answer as examples first. 
The example claims will be inside <!eC>...<eC!> tokens, evidence will be inside <!eE>...<eE!> tokens and the answer/reasoning will be inside <!eA>...<eA!> tokens. 
The answer is based on the evidence and it verifies whether the evidence supports or refutes the claim. 
For your task, you will be provided with climate-related claims and evidence in this format Q:[<!C> Claim: ...<C!> \textbackslash n <!E> Evidence: ...<E!>]. 
You will use the provided evidence to decide whether the associated claim is supported or refuted. 
You will first briefly explain your reasoning in one sentence, and then make the final judgement by writing LABEL: followed by a single word SUPPORTS or REFUTES. \\ \\

Here are some examples: \\
\{\texttt{examples}\} \\ \\

\hline

\end{tabular}
\caption{Prompts used in \climate.}
\label{tab:climate-prompts}
\end{table*}

%% file: tables/pheme_prompts.tex
\begin{table*}[!t]
\centering
\begin{tabular}{|p{2.0\columnwidth}|}
\hline

\textbf{ZS.} \\
You are an expert journalist. 
As an expert journalist, you will be helping us verify some rumours. 
For your task, you will be provided with rumours and evidence in this format Q:[<!R> Rumour: ...<R!> \textbackslash n <!E> Evidence: ...<E!>]. 
You will use the provided evidence to decide whether the associated rumour is supported or refuted. 
You will first briefly explain your reasoning in one sentence, and then make the final judgement by writing LABEL: followed by a single word SUPPORTS or REFUTES.  \\ \\

\textbf{CoT.} \\
You are an expert journalist. 
As an expert journalist, you will be helping us verify some rumours. 
For your task, you will be provided with rumours and evidence in this format Q:[<!R> Rumour: ...<R!> \textbackslash n <!E> Evidence: ...<E!>]. 
You will use the provided evidence to decide whether the associated rumour is supported or refuted. 
You will first briefly explain your reasoning in one sentence, and then make the final judgement by writing LABEL: followed by a single word SUPPORTS or REFUTES. 
Let's think step by step. \\ \\

\textbf{M-CoT.} \\
You are an expert journalist. 
As an expert journalist, you will be helping us verify some rumours. 
You will be provided with tuples of rumour, evidence and answer as examples first. 
The example rumours will be inside <!eR>...<eR!> tokens, evidence will be inside <!eE>...<eE!> tokens and the answer/reasoning will be inside <!eA>...<eA!> tokens. 
The answer is based on the evidence and it verifies whether the evidence supports or refutes the rumour. 
For your task, you will be provided with rumours and evidence in this format Q:[<!R> Rumour: ...<R!> \textbackslash n <!E> Evidence: ...<E!>]. 
You will use the provided evidence to decide whether the associated rumour is supported or refuted. 
You will first briefly explain your reasoning in one sentence, and then make the final judgement by writing LABEL: followed by a single word SUPPORTS or REFUTES. \\ \\

Here are some examples: \\
\{\texttt{examples}\} \\ \\

\hline

\end{tabular}
\caption{Prompts used in \pheme.}
\label{tab:pheme-prompts}
\end{table*}

%% file: tables/evaluation_vitc.tex
\begin{table*}[!t]
\begin{center}
\begin{adjustbox}{max width=\textwidth}
\small
\begin{tabular}{@{}l rr rr rr rr@{}}
\toprule
& \multicolumn{2}{c}{\textbf{EA} $\uparrow$} & \multicolumn{2}{c}{\textbf{FC} $\uparrow$} & \multicolumn{2}{c}{\textbf{BART} $\uparrow$} & \multicolumn{2}{c}{\textbf{PPL} $\downarrow$} \\
\midrule
\textbf{Model} & \textbf{Correct} & \textbf{Wrong} & \textbf{Correct} & \textbf{Wrong} & \textbf{Correct} & \textbf{Wrong} & \textbf{Correct} & \textbf{Wrong} \\
\midrule
Claude ZS & 0.86 & 0.75 & 0.86 & 0.77 & -4.15 & -4.21 & 103.49 & 75.94 \\
Claude ZS CoT & 0.86 & 0.61 & 0.86 & 0.75 & -4.42 & -4.19 & 55.44 & 39.25 \\
Claude M-CoT & 0.87 & 0.82 & 0.86 & 0.84 & -4.08 & -3.90 & 72.02 & 47.03 \\
GPT-4 ZS & 0.89 & 0.76 & 0.88 & 0.82 & -3.82 & -3.87 & 69.82 & 51.17 \\
GPT-4 ZS CoT & 0.89 & 0.76 & 0.88 & 0.80 & -3.77 & -3.8 & 61.13 & 47.85 \\
GPT-4 M-CoT & 0.90 & 0.77 & 0.90 & 0.75 & -2.98 & -3.01 & 46.39 & 40.95 \\
GPT-4o ZS & 0.91 & 0.79 & 0.90 & 0.81 & -3.67 & -3.41 & 57.60 & 24.46 \\
GPT-4o ZS CoT & 0.93 & 0.75 & 0.92 & 0.76 & -3.47 & -3.29 & 37.77 & 21.57 \\
GPT-4o M-CoT & 0.92 & 0.77 & 0.90 & 0.78 & -3.68 & -3.35 & 53.63 & 26.52 \\
\bottomrule
\end{tabular}
\end{adjustbox}
\end{center}
\caption{Qualitative evaluation on \vitc. We distinguish between correct and wrong claim veracity predictions.}
\label{tab:vitc_evaluation}
\end{table*}

%% file: tables/evaluation_climate.tex
\begin{table*}[!t]
\begin{center}
\begin{adjustbox}{max width=\textwidth}
\small
\begin{tabular}{@{}l rr rr rr rr@{}}
\toprule
& \multicolumn{2}{c}{\textbf{EA} $\uparrow$} & \multicolumn{2}{c}{\textbf{FC} $\uparrow$} & \multicolumn{2}{c}{\textbf{BART} $\uparrow$} & \multicolumn{2}{c}{\textbf{PPL} $\downarrow$} \\
\midrule
\textbf{Model} & \textbf{Correct} & \textbf{Wrong} & \textbf{Correct} & \textbf{Wrong} & \textbf{Correct} & \textbf{Wrong} & \textbf{Correct} & \textbf{Wrong} \\
\midrule
Claude ZS & 0.87 & 0.88 & 0.87 & 0.89 & -4.19 & -4.60 & 61.26 & 79.73 \\
Claude ZS CoT & 0.81 & 0.86 & 0.91 & 0.86 & -4.29 & -4.26 & 29.27 & 25.08 \\
Claude M-CoT & 0.90 & 0.82 & 0.91 & 0.85 & -3.36 & -3.68 & 33.51 & 33.33 \\
GPT-4 ZS & 0.93 & 0.79 & 0.93 & 0.84 & -3.65 & -3.79 & 30.17 & 36.24 \\
GPT-4 ZS CoT & 0.92 & 0.81 & 0.92 & 0.87 & -3.63 & -3.76 & 29.23 & 31.51 \\
GPT-4 M-CoT & 0.96 & 0.70 & 0.96 & 0.75 & -2.88 & -3.07 & 25.50 & 25.65 \\
GPT-4o ZS & 0.93 & 0.89 & 0.93 & 0.90 & -3.63 & -3.71 & 28.09 & 26.62 \\
GPT-4o ZS CoT & 0.95 & 0.85 & 0.95 & 0.90 & -3.37 & -3.48 & 20.77 & 19.95 \\
GPT-4o M-CoT & 0.90 & 0.88 & 0.91 & 0.86 & -3.56 & -3.72 & 28.15 & 28.04 \\
\bottomrule
\end{tabular}
\end{adjustbox}
\end{center}
\caption{Qualitative evaluation on \climate. We distinguish between correct and wrong claim veracity predictions.}
\label{tab:climate_evaluation}
\end{table*}

%% file: tables/evaluation_pheme.tex
\begin{table*}[!t]
\begin{center}
\begin{adjustbox}{max width=\textwidth}
\small
\begin{tabular}{@{}l rr rr rr rr@{}}
\toprule
& \multicolumn{2}{c}{\textbf{EA} $\uparrow$} & \multicolumn{2}{c}{\textbf{FC} $\uparrow$} & \multicolumn{2}{c}{\textbf{BART} $\uparrow$} & \multicolumn{2}{c}{\textbf{PPL} $\downarrow$} \\
\midrule
\textbf{Model} & \textbf{Correct} & \textbf{Wrong} & \textbf{Correct} & \textbf{Wrong} & \textbf{Correct} & \textbf{Wrong} & \textbf{Correct} & \textbf{Wrong} \\
\midrule
Claude ZS & 0.85 & 0.76 & 0.84 & 0.74 & -4.29 & -4.38 & 60.15 & 53.23 \\
Claude ZS CoT & 0.86 & 0.81 & 0.85 & 0.81 & -4.43 & -4.49 & 48.25 & 42.95 \\
Claude M-CoT & 0.89 & 0.87 & 0.89 & 0.86 & -4.11 & -4.30 & 57.30 & 58.42 \\
GPT-4 ZS & 0.88 & 0.81 & 0.89 & 0.81 & -3.90 & -3.87 & 41.33 & 35.52 \\
GPT-4 ZS CoT & 0.88 & 0.85 & 0.88 & 0.85 & -3.90 & -3.85 & 39.65 & 36.18 \\
GPT-4 M-CoT & 0.91 & 0.70 & 0.89 & 0.72 & -3.40 & -3.39 & 40.59 & 30.73 \\
GPT-4o ZS & 0.90 & 0.74 & 0.90 & 0.75 & -4.02 & -4.00 & 43.29 & 34.20 \\
GPT-4o ZS CoT & 0.92 & 0.82 & 0.92 & 0.84 & -3.76 & -3.70 & 30.33 & 24.55 \\
GPT-4o M-CoT & 0.89 & 0.79 & 0.89 & 0.79 & -4.08 & -4.05 & 49.46 & 40.59 \\
\bottomrule
\end{tabular}
\end{adjustbox}
\end{center}
\caption{Qualitative evaluation on \pheme. We distinguish between correct and wrong claim veracity predictions.}
\label{tab:pheme_evaluation}
\end{table*}

%% file: tables/permutation_test_vitc.tex
\begin{table*}[!t]
\begin{center}
\begin{adjustbox}{max width=\textwidth}
\begin{tabular}{@{}lrrrrrrrr@{}}
\toprule
 & \textbf{Claude ZS} & \textbf{Claude ZS CoT} & \textbf{Claude M-CoT} & \textbf{GPT-4 ZS} & \textbf{GPT-4 ZS CoT} & \textbf{GPT-4 M-CoT} & \textbf{GPT-4o ZS} & \textbf{GPT-4o ZS CoT} \\
\midrule
\textbf{Claude ZS CoT} & 0.4061 & - & - & - & - & - & - & - \\
\textbf{Claude M-CoT} & 0.5985 & 0.1768 & - & - & - & - & - & - \\
\textbf{GPT-4 ZS} & 0.3461 & 0.0639 & 0.6466 & - & - & - & - & - \\
\textbf{GPT-4 ZS CoT} & 0.3303 & 0.0600 & 0.6636 & 0.9703 & - & - & - & - \\
\textbf{GPT-4 M-CoT} & 0.0830 & 0.0066 & 0.1943 & 0.3769 & 0.3489 & - & - & - \\
\textbf{GPT-4o ZS} & 0.0732 & 0.0067 & 0.1897 & 0.4086 & 0.3739 & 0.8896 & - & - \\
\textbf{GPT-4o ZS CoT} & 0.0158 & 0.0002 & 0.0558 & 0.1511 & 0.1269 & 0.6856 & 0.5457 & - \\
\textbf{GPT-4o M-CoT} & 0.0866 & 0.0064 & 0.2211 & 0.4259 & 0.3837 & 0.8893 & 0.9988 & 0.5626 \\
\bottomrule
\end{tabular}
\end{adjustbox}
\end{center}
\caption{Pairwise Permutation Test on 100 evaluation samples from \vitc.} 
\label{tab:pairwise_permutation_vitc}
\end{table*}

%% file: tables/permutation_test_climate.tex
\begin{table*}[!t]
\begin{center}
\begin{adjustbox}{max width=\textwidth}
\begin{tabular}{@{}lrrrrrrrr@{}}
\toprule
& \textbf{Claude ZS} & \textbf{Claude ZS CoT} & \textbf{Claude M-CoT} & \textbf{GPT-4 ZS} & \textbf{GPT-4 ZS CoT} & \textbf{GPT-4 M-CoT} & \textbf{GPT-4o ZS} & \textbf{GPT-4o ZS CoT} \\
\midrule
\textbf{Claude ZS CoT} & 0.0480 & - & - & - & - & - & - & -  \\
\textbf{Claude M-CoT} & 0.4062 & 0.0027 & - & - & - & - & - & -  \\
\textbf{GPT-4 ZS} & 0.0811 & 0.0001 & 0.5795 & - & - & - & - & - \\
\textbf{GPT-4 ZS CoT} & 0.0889 & 0.0002 & 0.6516 & 0.8770 & - & - & - & - \\
\textbf{GPT-4 M-CoT} & 0.0022 & 0.0001 & 0.0802 & 0.1529 & 0.1005 & - & - & - \\
\textbf{GPT-4o ZS} & 0.0108 & 0.0001 & 0.2442 & 0.4047 & 0.3009 & 0.5190 & - & - \\
\textbf{GPT-4o ZS CoT} & 0.0006 & 0.0001 & 0.0528 & 0.0932 & 0.0536 & 0.9319 & 0.4036 & - \\
\textbf{GPT-4o M-CoT} & 0.1325 & 0.0001 & 0.7312 & 0.7618 & 0.8751 & 0.0741 & 0.2653 & 0.0369 \\
\bottomrule
\end{tabular}
\end{adjustbox}
\end{center}
\caption{Pairwise Permutation Test on 100 evaluation samples from \climate.} 
\label{tab:pairwise_permutation_climate}
\end{table*}

%% file: tables/permutation_test_pheme.tex
\begin{table*}[!ht]
\begin{center}
\begin{adjustbox}{max width=\textwidth}
\begin{tabular}{@{}lrrrrrrrr@{}}
\toprule
& \textbf{Claude ZS} & \textbf{Claude ZS CoT} & \textbf{Claude M-CoT} & \textbf{GPT-4 ZS} & \textbf{GPT-4 ZS CoT} & \textbf{GPT-4 M-CoT} & \textbf{GPT-4o ZS} & \textbf{GPT-4o ZS CoT} \\
\midrule
\textbf{Claude ZS CoT} & 0.1872 & - & - & - & - & - & - & - \\
\textbf{Claude M-CoT} & 0.0068 & 0.0953 & - & - & - & - & - & - \\
\textbf{GPT-4 ZS} & 0.0276 & 0.3199 & 0.4636 & - & - & - & - & - \\
\textbf{GPT-4 ZS CoT} & 0.0082 & 0.1534 & 0.6888 & 0.7094 & - & - & - & - \\
\textbf{GPT-4 M-CoT} & 0.2363 & 0.8827 & 0.2256 & 0.4959 & 0.3116 & - & - & - \\
\textbf{GPT-4o ZS} & 0.0534 & 0.4611 & 0.3528 & 0.8139 & 0.5373 & 0.6487 & - & - \\
\textbf{GPT-4o ZS CoT} & 0.0002 & 0.0033 & 0.4062 & 0.0791 & 0.1584 & 0.0263 & 0.0518 & - \\
\textbf{GPT-4o M-CoT} & 0.0270 & 0.3188 & 0.4208 & 0.9707 & 0.6609 & 0.5043 & 0.8387 & 0.0681 \\
\bottomrule
\end{tabular}
\end{adjustbox}
\end{center}
\caption{Pairwise Permutation Test on 100 evaluation samples from \pheme.}
\label{tab:pairwise_permutation_pheme}
\end{table*}

%% file: main_v2.bbl
\begin{thebibliography}{92}
\expandafter\ifx\csname natexlab\endcsname\relax\def\natexlab#1{#1}\fi

\bibitem[{Ahrabian et~al.(2024)Ahrabian, Sourati, Sun, Zhang, Jiang, Morstatter, and Pujara}]{ahrabian-etal-2024-curious}
Kian Ahrabian, Zhivar Sourati, Kexuan Sun, Jiarui Zhang, Yifan Jiang, Fred Morstatter, and Jay Pujara. 2024.
\newblock \href {https://doi.org/10.48550/ARXIV.2401.12117} {The curious case of nonverbal abstract reasoning with multi-modal large language models}.
\newblock \emph{CoRR}, abs/2401.12117.

\bibitem[{Akyürek et~al.(2024)Akyürek, Akyürek, Choshen, Wijaya, and Andreas}]{akyürek2024deductive}
Afra~Feyza Akyürek, Ekin Akyürek, Leshem Choshen, Derry Wijaya, and Jacob Andreas. 2024.
\newblock \href {http://arxiv.org/abs/2401.08574} {Deductive closure training of language models for coherence, accuracy, and updatability}.

\bibitem[{Alhindi et~al.(2023)Alhindi, Muresan, and Nakov}]{alhindi2023large}
Tariq Alhindi, Smaranda Muresan, and Preslav Nakov. 2023.
\newblock \href {http://arxiv.org/abs/2311.09552} {Large language models are few-shot training example generators: A case study in fallacy recognition}.

\bibitem[{Aly et~al.(2023)Aly, Strong, and Vlachos}]{aly-etal-2023-qa}
Rami Aly, Marek Strong, and Andreas Vlachos. 2023.
\newblock \href {https://doi.org/10.18653/v1/2023.emnlp-main.521} {{QA}-{N}at{V}er: Question answering for natural logic-based fact verification}.
\newblock In \emph{Proceedings of the 2023 Conference on Empirical Methods in Natural Language Processing}, pages 8376--8391, Singapore. Association for Computational Linguistics.

\bibitem[{Anthropic(2023)}]{anthropic-2023-claude-v3}
Anthropic. 2023.
\newblock \href {https://www.anthropic.com/news/claude-3-family} {The claude 3 model family: Opus, sonnet, haiku}.

\bibitem[{Bakhtin et~al.(2022)Bakhtin, Brown, Dinan, Farina, Flaherty, Fried, Goff, Gray, Hu, Jacob, Komeili, Konath, Kwon, Lerer, Lewis, Miller, Mitts, Renduchintala, Roller, Rowe, Shi, Spisak, Wei, Wu, Zhang, and Zijlstra}]{fair2022}
Anton Bakhtin, Noam Brown, Emily Dinan, Gabriele Farina, Colin Flaherty, Daniel Fried, Andrew Goff, Jonathan Gray, Hengyuan Hu, Athul~Paul Jacob, Mojtaba Komeili, Karthik Konath, Minae Kwon, Adam Lerer, Mike Lewis, Alexander~H. Miller, Sasha Mitts, Adithya Renduchintala, Stephen Roller, Dirk Rowe, Weiyan Shi, Joe Spisak, Alexander Wei, David Wu, Hugh Zhang, and Markus Zijlstra. 2022.
\newblock \href {https://doi.org/10.1126/science.ade9097} {Human-level play in the game of diplomacy by combining language models with strategic reasoning}.
\newblock \emph{Science}, 378(6624):1067--1074.

\bibitem[{Bennet et~al.(1954)Bennet, Alpert, and Goldstein}]{10.1086/266520}
E.~M. Bennet, R.~Alpert, and A.~C. Goldstein. 1954.
\newblock \href {https://doi.org/10.1086/266520} {Communications through limited-response questioning*}.
\newblock \emph{Public Opinion Quarterly}, 18(3):303--308.

\bibitem[{Bouyamourn(2023)}]{bouyamourn-2023-llms}
Adam Bouyamourn. 2023.
\newblock \href {https://doi.org/10.18653/v1/2023.emnlp-main.192} {Why {LLM}s hallucinate, and how to get (evidential) closure: Perceptual, intensional, and extensional learning for faithful natural language generation}.
\newblock In \emph{Proceedings of the 2023 Conference on Empirical Methods in Natural Language Processing}, pages 3181--3193, Singapore. Association for Computational Linguistics.

\bibitem[{Bubeck et~al.(2023)Bubeck, Chandrasekaran, Eldan, Gehrke, Horvitz, Kamar, Lee, Lee, Li, Lundberg, Nori, Palangi, Ribeiro, and Zhang}]{bubeck2023sparks}
Sébastien Bubeck, Varun Chandrasekaran, Ronen Eldan, Johannes Gehrke, Eric Horvitz, Ece Kamar, Peter Lee, Yin~Tat Lee, Yuanzhi Li, Scott Lundberg, Harsha Nori, Hamid Palangi, Marco~Tulio Ribeiro, and Yi~Zhang. 2023.
\newblock \href {http://arxiv.org/abs/2303.12712} {Sparks of artificial general intelligence: Early experiments with gpt-4}.

\bibitem[{Cao(2023)}]{cao2023enhancing}
Lang Cao. 2023.
\newblock \href {http://arxiv.org/abs/2308.09267} {Enhancing reasoning capabilities of large language models: A graph-based verification approach}.

\bibitem[{Chang and Bergen(2024)}]{chang-etal-2024-language}
Tyler~A. Chang and Benjamin~K. Bergen. 2024.
\newblock \href {https://doi.org/10.1162/coli_a_00492} {Language model behavior: A comprehensive survey}.
\newblock \emph{Computational Linguistics}, 50(1):293--350.

\bibitem[{Chen et~al.(2024)Chen, Chi, Wang, and Zhou}]{chen-etal-2024-premise}
Xinyun Chen, Ryan~A. Chi, Xuezhi Wang, and Denny Zhou. 2024.
\newblock Premise order matters in reasoning with large language models.
\newblock In \emph{Proceedings of the 41st International Conference on Machine Learning}, ICML'24. JMLR.org.

\bibitem[{Cheung and Lam(2023)}]{factllama}
Tsun-Hin Cheung and Kin-Man Lam. 2023.
\newblock \href {http://arxiv.org/abs/2309.00240} {Factllama: Optimizing instruction-following language models with external knowledge for automated fact-checking}.

\bibitem[{Chollet(2023)}]{substackThinkAbout}
François Chollet. 2023.
\newblock {H}ow {I} think about {L}{L}{M} prompt engineering --- fchollet.substack.com.
\newblock \url{https://fchollet.substack.com/p/how-i-think-about-llm-prompt-engineering}.
\newblock [Accessed 11-02-2025].

\bibitem[{DeepSeek-AI et~al.(2025)DeepSeek-AI, Guo, Yang, Zhang, Song, Zhang, Xu, Zhu, Ma, Wang, Bi, Zhang, Yu, Wu, Wu, Gou, Shao, Li, Gao, Liu, Xue, Wang, Wu, Feng, Lu, Zhao, Deng, Zhang, Ruan, Dai, Chen, Ji, Li, Lin, Dai, Luo, Hao, Chen, Li, Zhang, Bao, Xu, Wang, Ding, Xin, Gao, Qu, Li, Guo, Li, Wang, Chen, Yuan, Qiu, Li, Cai, Ni, Liang, Chen, Dong, Hu, Gao, Guan, Huang, Yu, Wang, Zhang, Zhao, Wang, Zhang, Xu, Xia, Zhang, Zhang, Tang, Li, Wang, Li, Tian, Huang, Zhang, Wang, Chen, Du, Ge, Zhang, Pan, Wang, Chen, Jin, Chen, Lu, Zhou, Chen, Ye, Wang, Yu, Zhou, Pan, Li, Zhou, Wu, Ye, Yun, Pei, Sun, Wang, Zeng, Zhao, Liu, Liang, Gao, Yu, Zhang, Xiao, An, Liu, Wang, Chen, Nie, Cheng, Liu, Xie, Liu, Yang, Li, Su, Lin, Li, Jin, Shen, Chen, Sun, Wang, Song, Zhou, Wang, Shan, Li, Wang, Wei, Zhang, Xu, Li, Zhao, Sun, Wang, Yu, Zhang, Shi, Xiong, He, Piao, Wang, Tan, Ma, Liu, Guo, Ou, Wang, Gong, Zou, He, Xiong, Luo, You, Liu, Zhou, Zhu, Xu, Huang, Li, Zheng, Zhu, Ma, Tang, Zha, Yan, Ren, Ren, Sha, Fu, Xu, Xie, Zhang,
  Hao, Ma, Yan, Wu, Gu, Zhu, Liu, Li, Xie, Song, Pan, Huang, Xu, Zhang, and Zhang}]{deepseekai2025deepseekr1incentivizingreasoningcapability}
DeepSeek-AI, Daya Guo, Dejian Yang, Haowei Zhang, Junxiao Song, Ruoyu Zhang, Runxin Xu, Qihao Zhu, Shirong Ma, Peiyi Wang, Xiao Bi, Xiaokang Zhang, Xingkai Yu, Yu~Wu, Z.~F. Wu, Zhibin Gou, Zhihong Shao, Zhuoshu Li, Ziyi Gao, Aixin Liu, Bing Xue, Bingxuan Wang, Bochao Wu, Bei Feng, Chengda Lu, Chenggang Zhao, Chengqi Deng, Chenyu Zhang, Chong Ruan, Damai Dai, Deli Chen, Dongjie Ji, Erhang Li, Fangyun Lin, Fucong Dai, Fuli Luo, Guangbo Hao, Guanting Chen, Guowei Li, H.~Zhang, Han Bao, Hanwei Xu, Haocheng Wang, Honghui Ding, Huajian Xin, Huazuo Gao, Hui Qu, Hui Li, Jianzhong Guo, Jiashi Li, Jiawei Wang, Jingchang Chen, Jingyang Yuan, Junjie Qiu, Junlong Li, J.~L. Cai, Jiaqi Ni, Jian Liang, Jin Chen, Kai Dong, Kai Hu, Kaige Gao, Kang Guan, Kexin Huang, Kuai Yu, Lean Wang, Lecong Zhang, Liang Zhao, Litong Wang, Liyue Zhang, Lei Xu, Leyi Xia, Mingchuan Zhang, Minghua Zhang, Minghui Tang, Meng Li, Miaojun Wang, Mingming Li, Ning Tian, Panpan Huang, Peng Zhang, Qiancheng Wang, Qinyu Chen, Qiushi Du, Ruiqi Ge, Ruisong
  Zhang, Ruizhe Pan, Runji Wang, R.~J. Chen, R.~L. Jin, Ruyi Chen, Shanghao Lu, Shangyan Zhou, Shanhuang Chen, Shengfeng Ye, Shiyu Wang, Shuiping Yu, Shunfeng Zhou, Shuting Pan, S.~S. Li, Shuang Zhou, Shaoqing Wu, Shengfeng Ye, Tao Yun, Tian Pei, Tianyu Sun, T.~Wang, Wangding Zeng, Wanjia Zhao, Wen Liu, Wenfeng Liang, Wenjun Gao, Wenqin Yu, Wentao Zhang, W.~L. Xiao, Wei An, Xiaodong Liu, Xiaohan Wang, Xiaokang Chen, Xiaotao Nie, Xin Cheng, Xin Liu, Xin Xie, Xingchao Liu, Xinyu Yang, Xinyuan Li, Xuecheng Su, Xuheng Lin, X.~Q. Li, Xiangyue Jin, Xiaojin Shen, Xiaosha Chen, Xiaowen Sun, Xiaoxiang Wang, Xinnan Song, Xinyi Zhou, Xianzu Wang, Xinxia Shan, Y.~K. Li, Y.~Q. Wang, Y.~X. Wei, Yang Zhang, Yanhong Xu, Yao Li, Yao Zhao, Yaofeng Sun, Yaohui Wang, Yi~Yu, Yichao Zhang, Yifan Shi, Yiliang Xiong, Ying He, Yishi Piao, Yisong Wang, Yixuan Tan, Yiyang Ma, Yiyuan Liu, Yongqiang Guo, Yuan Ou, Yuduan Wang, Yue Gong, Yuheng Zou, Yujia He, Yunfan Xiong, Yuxiang Luo, Yuxiang You, Yuxuan Liu, Yuyang Zhou, Y.~X. Zhu,
  Yanhong Xu, Yanping Huang, Yaohui Li, Yi~Zheng, Yuchen Zhu, Yunxian Ma, Ying Tang, Yukun Zha, Yuting Yan, Z.~Z. Ren, Zehui Ren, Zhangli Sha, Zhe Fu, Zhean Xu, Zhenda Xie, Zhengyan Zhang, Zhewen Hao, Zhicheng Ma, Zhigang Yan, Zhiyu Wu, Zihui Gu, Zijia Zhu, Zijun Liu, Zilin Li, Ziwei Xie, Ziyang Song, Zizheng Pan, Zhen Huang, Zhipeng Xu, Zhongyu Zhang, and Zhen Zhang. 2025.
\newblock \href {http://arxiv.org/abs/2501.12948} {Deepseek-r1: Incentivizing reasoning capability in llms via reinforcement learning}.

\bibitem[{Diggelmann et~al.(2020)Diggelmann, Boyd-Graber, Bulian, Ciaramita, and Leippold}]{diggelmann2020climatefever}
Thomas Diggelmann, Jordan Boyd-Graber, Jannis Bulian, Massimiliano Ciaramita, and Markus Leippold. 2020.
\newblock \href {http://arxiv.org/abs/2012.00614} {Climate-fever: A dataset for verification of real-world climate claims}.

\bibitem[{Dougrez-Lewis et~al.(2022)Dougrez-Lewis, Kochkina, Arana-Catania, Liakata, and He}]{phemeplus}
John Dougrez-Lewis, Elena Kochkina, Miguel Arana-Catania, Maria Liakata, and Yulan He. 2022.
\newblock \href {https://doi.org/10.18653/v1/2022.fever-1.6} {{PHEMEP}lus: Enriching social media rumour verification with external evidence}.
\newblock In \emph{Proceedings of the Fifth Fact Extraction and VERification Workshop (FEVER)}, pages 49--58, Dublin, Ireland. Association for Computational Linguistics.

\bibitem[{Dutta et~al.(2024)Dutta, Singh, Chakrabarti, and Chakraborty}]{Dutta2024HowTT}
Subhabrata Dutta, Joykirat Singh, Soumen Chakrabarti, and Tanmoy Chakraborty. 2024.
\newblock \href {https://api.semanticscholar.org/CorpusID:268041831} {How to think step-by-step: A mechanistic understanding of chain-of-thought reasoning}.
\newblock \emph{ArXiv}, abs/2402.18312.

\bibitem[{Flach and Kakas(2000)}]{Flach2000}
Peter~A. Flach and Antonis~C. Kakas. 2000.
\newblock \href {https://doi.org/10.1007/978-94-017-0606-3_1} {\emph{Abductive and Inductive Reasoning: Background and Issues}}, page 1–27. Springer Netherlands.

\bibitem[{Galotti(1989)}]{Galotti1989ApproachesTS}
Kathleen~M. Galotti. 1989.
\newblock \href {https://api.semanticscholar.org/CorpusID:143779846} {Approaches to studying formal and everyday reasoning.}
\newblock \emph{Psychological Bulletin}, 105:331--351.

\bibitem[{Gandhi et~al.(2023)Gandhi, Sadigh, and Goodman}]{Gandhi2023StrategicRW}
Kanishk Gandhi, Dorsa Sadigh, and Noah~D. Goodman. 2023.
\newblock \href {https://api.semanticscholar.org/CorpusID:258968043} {Strategic reasoning with language models}.
\newblock \emph{ArXiv}, abs/2305.19165.

\bibitem[{Gao et~al.(2023)Gao, Ding, Qin, and Liu}]{gao-etal-2023-chatgpt}
Jinglong Gao, Xiao Ding, Bing Qin, and Ting Liu. 2023.
\newblock \href {https://doi.org/10.18653/v1/2023.findings-emnlp.743} {Is {C}hat{GPT} a good causal reasoner? a comprehensive evaluation}.
\newblock In \emph{Findings of the Association for Computational Linguistics: EMNLP 2023}, pages 11111--11126, Singapore. Association for Computational Linguistics.

\bibitem[{Guo et~al.(2023)Guo, Zhang, Wang, Jiang, Nie, Ding, Yue, and Wu}]{guo2023close}
Biyang Guo, Xin Zhang, Ziyuan Wang, Minqi Jiang, Jinran Nie, Yuxuan Ding, Jianwei Yue, and Yupeng Wu. 2023.
\newblock \href {http://arxiv.org/abs/2301.07597} {How close is chatgpt to human experts? comparison corpus, evaluation, and detection}.

\bibitem[{He et~al.(2025)He, Dong, Ding, and Li}]{he2025towards}
Shwai He, Daize Dong, Liang Ding, and Ang Li. 2025.
\newblock \href {https://openreview.net/forum?id=HTpMOl6xSI} {Towards efficient mixture of experts: A holistic study of compression techniques}.
\newblock \emph{Transactions on Machine Learning Research}.

\bibitem[{Hosseini et~al.(2024)Hosseini, Sordoni, Toyama, Courville, and Agarwal}]{hosseini-etal-2024-not}
Arian Hosseini, Alessandro Sordoni, Daniel~Kenji Toyama, Aaron Courville, and Rishabh Agarwal. 2024.
\newblock \href {https://openreview.net/forum?id=RcqAmkDJfI} {Not all {LLM} reasoners are created equal}.
\newblock In \emph{The 4th Workshop on Mathematical Reasoning and AI at NeurIPS'24}.

\bibitem[{Hu et~al.(2023{\natexlab{a}})Hu, Storks, Lewis, and Chai}]{hu-etal-2023-context}
Xiaoyang Hu, Shane Storks, Richard Lewis, and Joyce Chai. 2023{\natexlab{a}}.
\newblock \href {https://doi.org/10.18653/v1/2023.acl-long.109} {In-context analogical reasoning with pre-trained language models}.
\newblock In \emph{Proceedings of the 61st Annual Meeting of the Association for Computational Linguistics (Volume 1: Long Papers)}, pages 1953--1969, Toronto, Canada. Association for Computational Linguistics.

\bibitem[{Hu et~al.(2023{\natexlab{b}})Hu, Chen, Li, Guo, Wen, Yu, and Guo}]{hu-etal-2023-large}
Xuming Hu, Junzhe Chen, Xiaochuan Li, Yufei Guo, Lijie Wen, Philip~S. Yu, and Zhijiang Guo. 2023{\natexlab{b}}.
\newblock \href {https://doi.org/10.48550/ARXIV.2310.05177} {Do large language models know about facts?}
\newblock \emph{CoRR}, abs/2310.05177.

\bibitem[{Huang and Chang(2023)}]{huang-chang-2023-towards}
Jie Huang and Kevin Chen-Chuan Chang. 2023.
\newblock \href {https://doi.org/10.18653/v1/2023.findings-acl.67} {Towards reasoning in large language models: A survey}.
\newblock In \emph{Findings of the Association for Computational Linguistics: ACL 2023}, pages 1049--1065, Toronto, Canada. Association for Computational Linguistics.

\bibitem[{Kavumba et~al.(2019)Kavumba, Inoue, Heinzerling, Singh, Reisert, and Inui}]{kavumba-etal-2019-choosing}
Pride Kavumba, Naoya Inoue, Benjamin Heinzerling, Keshav Singh, Paul Reisert, and Kentaro Inui. 2019.
\newblock \href {https://doi.org/10.18653/v1/D19-6004} {When choosing plausible alternatives, clever hans can be clever}.
\newblock In \emph{Proceedings of the First Workshop on Commonsense Inference in Natural Language Processing}, pages 33--42, Hong Kong, China. Association for Computational Linguistics.

\bibitem[{Kojima et~al.(2023)Kojima, Gu, Reid, Matsuo, and Iwasawa}]{kojima2023large}
Takeshi Kojima, Shixiang~Shane Gu, Machel Reid, Yutaka Matsuo, and Yusuke Iwasawa. 2023.
\newblock \href {http://arxiv.org/abs/2205.11916} {Large language models are zero-shot reasoners}.

\bibitem[{Kosinski(2023)}]{kosinski2023theory}
Michal Kosinski. 2023.
\newblock \href {http://arxiv.org/abs/2302.02083} {Theory of mind might have spontaneously emerged in large language models}.

\bibitem[{Kung et~al.(2023)Kung, Cheatham, Medenilla, Sillos, De~Leon, Elepaño, Madriaga, Aggabao, Diaz-Candido, Maningo, and Tseng}]{Kung2023}
Tiffany~H. Kung, Morgan Cheatham, Arielle Medenilla, Czarina Sillos, Lorie De~Leon, Camille Elepaño, Maria Madriaga, Rimel Aggabao, Giezel Diaz-Candido, James Maningo, and Victor Tseng. 2023.
\newblock \href {https://doi.org/10.1371/journal.pdig.0000198} {Performance of chatgpt on usmle: Potential for ai-assisted medical education using large language models}.
\newblock \emph{PLOS Digital Health}, 2(2):e0000198.

\bibitem[{Lee et~al.(2020)Lee, Li, Wang, Yih, Ma, and Khabsa}]{lee2020language}
Nayeon Lee, Belinda~Z. Li, Sinong Wang, Wen-tau Yih, Hao Ma, and Madian Khabsa. 2020.
\newblock \href {https://doi.org/10.18653/v1/2020.fever-1.5} {Language models as fact checkers?}

\bibitem[{Lewis and Mitchell(2024)}]{lewis-etal-2024-counterfactual}
Martha Lewis and Melanie Mitchell. 2024.
\newblock \href {https://doi.org/10.48550/ARXIV.2402.08955} {Using counterfactual tasks to evaluate the generality of analogical reasoning in large language models}.
\newblock \emph{CoRR}, abs/2402.08955.

\bibitem[{Lewkowycz et~al.(2022)Lewkowycz, Andreassen, Dohan, Dyer, Michalewski, Ramasesh, Slone, Anil, Schlag, Gutman-Solo, Wu, Neyshabur, Gur-Ari, and Misra}]{lewkowycz2022solving}
Aitor Lewkowycz, Anders Andreassen, David Dohan, Ethan Dyer, Henryk Michalewski, Vinay Ramasesh, Ambrose Slone, Cem Anil, Imanol Schlag, Theo Gutman-Solo, Yuhuai Wu, Behnam Neyshabur, Guy Gur-Ari, and Vedant Misra. 2022.
\newblock \href {http://arxiv.org/abs/2206.14858} {Solving quantitative reasoning problems with language models}.

\bibitem[{Li et~al.(2023{\natexlab{a}})Li, Peng, and Zhang}]{li2023selfchecker}
Miaoran Li, Baolin Peng, and Zhu Zhang. 2023{\natexlab{a}}.
\newblock \href {http://arxiv.org/abs/2305.14623} {Self-checker: Plug-and-play modules for fact-checking with large language models}.

\bibitem[{Li et~al.(2024)Li, Tian, Jiao, Chen, Zhao, and Jiang}]{li-etal-2024-look}
Yian Li, Wentao Tian, Yang Jiao, Jingjing Chen, Na~Zhao, and Yu-Gang Jiang. 2024.
\newblock \href {http://arxiv.org/abs/2404.12966} {Look before you decide: Prompting active deduction of mllms for assumptive reasoning}.
\newblock \emph{arXiv preprint arXiv:2404.12966}.

\bibitem[{Li et~al.(2023{\natexlab{b}})Li, Lin, Zhang, Fu, Chen, Lou, and Chen}]{li2023making}
Yifei Li, Zeqi Lin, Shizhuo Zhang, Qiang Fu, Bei Chen, Jian-Guang Lou, and Weizhu Chen. 2023{\natexlab{b}}.
\newblock \href {https://doi.org/10.18653/v1/2023.acl-long.291} {Making language models better reasoners with step-aware verifier}.
\newblock In \emph{Proceedings of the 61st Annual Meeting of the Association for Computational Linguistics (Volume 1: Long Papers)}, pages 5315--5333, Toronto, Canada. Association for Computational Linguistics.

\bibitem[{Lin et~al.(2023)Lin, Luo, Ma, and Chen}]{lin-etal-2023-beneath}
Hongzhan Lin, Ziyang Luo, Jing Ma, and Long Chen. 2023.
\newblock \href {https://doi.org/10.18653/v1/2023.findings-emnlp.611} {Beneath the surface: Unveiling harmful memes with multimodal reasoning distilled from large language models}.
\newblock In \emph{Findings of the Association for Computational Linguistics: EMNLP 2023}, pages 9114--9128, Singapore. Association for Computational Linguistics.

\bibitem[{Lu et~al.(2023{\natexlab{a}})Lu, Peng, Cheng, Galley, Chang, Wu, Zhu, and Gao}]{lu2023chameleon}
Pan Lu, Baolin Peng, Hao Cheng, Michel Galley, Kai-Wei Chang, Ying~Nian Wu, Song-Chun Zhu, and Jianfeng Gao. 2023{\natexlab{a}}.
\newblock Chameleon: Plug-and-play compositional reasoning with large language models.
\newblock \emph{arXiv preprint arXiv:2304.09842}.

\bibitem[{Lu et~al.(2023{\natexlab{b}})Lu, Bigoulaeva, Sachdeva, Madabushi, and Gurevych}]{lu2023emergent}
Sheng Lu, Irina Bigoulaeva, Rachneet Sachdeva, Harish~Tayyar Madabushi, and Iryna Gurevych. 2023{\natexlab{b}}.
\newblock \href {http://arxiv.org/abs/2309.01809} {Are emergent abilities in large language models just in-context learning?}

\bibitem[{McKnight and Najab(2010)}]{mcknight2010mann}
Patrick~E McKnight and Julius Najab. 2010.
\newblock Mann-whitney u test.
\newblock \emph{The Corsini encyclopedia of psychology}, pages 1--1.

\bibitem[{Morishita et~al.(2023)Morishita, Morio, Yamaguchi, and Sogawa}]{pmlr-v202-morishita23a}
Terufumi Morishita, Gaku Morio, Atsuki Yamaguchi, and Yasuhiro Sogawa. 2023.
\newblock \href {https://proceedings.mlr.press/v202/morishita23a.html} {Learning deductive reasoning from synthetic corpus based on formal logic}.
\newblock In \emph{Proceedings of the 40th International Conference on Machine Learning}, volume 202 of \emph{Proceedings of Machine Learning Research}, pages 25254--25274. PMLR.

\bibitem[{Mündler et~al.(2023)Mündler, He, Jenko, and Vechev}]{mündler2023selfcontradictory}
Niels Mündler, Jingxuan He, Slobodan Jenko, and Martin Vechev. 2023.
\newblock \href {http://arxiv.org/abs/2305.15852} {Self-contradictory hallucinations of large language models: Evaluation, detection and mitigation}.

\bibitem[{Nori et~al.(2023)Nori, King, McKinney, Carignan, and Horvitz}]{nori2023capabilities}
Harsha Nori, Nicholas King, Scott~Mayer McKinney, Dean Carignan, and Eric Horvitz. 2023.
\newblock \href {http://arxiv.org/abs/2303.13375} {Capabilities of gpt-4 on medical challenge problems}.

\bibitem[{OpenAI et~al.(2024)OpenAI, :, Hurst, Lerer, Goucher, Perelman, Ramesh, Clark, Ostrow, Welihinda, Hayes, Radford, Mądry, Baker-Whitcomb, Beutel, Borzunov, Carney, Chow, Kirillov, Nichol, Paino, Renzin, Passos, Kirillov, Christakis, Conneau, Kamali, Jabri, Moyer, Tam, Crookes, Tootoochian, Tootoonchian, Kumar, Vallone, Karpathy, Braunstein, Cann, Codispoti, Galu, Kondrich, Tulloch, Mishchenko, Baek, Jiang, Pelisse, Woodford, Gosalia, Dhar, Pantuliano, Nayak, Oliver, Zoph, Ghorbani, Leimberger, Rossen, Sokolowsky, Wang, Zweig, Hoover, Samic, McGrew, Spero, Giertler, Cheng, Lightcap, Walkin, Quinn, Guarraci, Hsu, Kellogg, Eastman, Lugaresi, Wainwright, Bassin, Hudson, Chu, Nelson, Li, Shern, Conger, Barette, Voss, Ding, Lu, Zhang, Beaumont, Hallacy, Koch, Gibson, Kim, Choi, McLeavey, Hesse, Fischer, Winter, Czarnecki, Jarvis, Wei, Koumouzelis, Sherburn, Kappler, Levin, Levy, Carr, Farhi, Mely, Robinson, Sasaki, Jin, Valladares, Tsipras, Li, Nguyen, Findlay, Oiwoh, Wong, Asdar, Proehl, Yang, Antonow,
  Kramer, Peterson, Sigler, Wallace, Brevdo, Mays, Khorasani, Such, Raso, Zhang, von Lohmann, Sulit, Goh, Oden, Salmon, Starace, Brockman, Salman, Bao, Hu, Wong, Wang, Schmidt, Whitney, Jun, Kirchner, de~Oliveira~Pinto, Ren, Chang, Chung, Kivlichan, O'Connell, O'Connell, Osband, Silber, Sohl, Okuyucu, Lan, Kostrikov, Sutskever, Kanitscheider, Gulrajani, Coxon, Menick, Pachocki, Aung, Betker, Crooks, Lennon, Kiros, Leike, Park, Kwon, Phang, Teplitz, Wei, Wolfe, Chen, Harris, Varavva, Lee, Shieh, Lin, Yu, Weng, Tang, Yu, Jang, Candela, Beutler, Landers, Parish, Heidecke, Schulman, Lachman, McKay, Uesato, Ward, Kim, Huizinga, Sitkin, Kraaijeveld, Gross, Kaplan, Snyder, Achiam, Jiao, Lee, Zhuang, Harriman, Fricke, Hayashi, Singhal, Shi, Karthik, Wood, Rimbach, Hsu, Nguyen, Gu-Lemberg, Button, Liu, Howe, Muthukumar, Luther, Ahmad, Kai, Itow, Workman, Pathak, Chen, Jing, Guy, Fedus, Zhou, Mamitsuka, Weng, McCallum, Held, Ouyang, Feuvrier, Zhang, Kondraciuk, Kaiser, Hewitt, Metz, Doshi, Aflak, Simens, Boyd,
  Thompson, Dukhan, Chen, Gray, Hudnall, Zhang, Aljubeh, Litwin, Zeng, Johnson, Shetty, Gupta, Shah, Yatbaz, Yang, Zhong, Glaese, Chen, Janner, Lampe, Petrov, Wu, Wang, Fradin, Pokrass, Castro, de~Castro, Pavlov, Brundage, Wang, Khan, Murati, Bavarian, Lin, Yesildal, Soto, Gimelshein, Cone, Staudacher, Summers, LaFontaine, Chowdhury, Ryder, Stathas, Turley, Tezak, Felix, Kudige, Keskar, Deutsch, Bundick, Puckett, Nachum, Okelola, Boiko, Murk, Jaffe, Watkins, Godement, Campbell-Moore, Chao, McMillan, Belov, Su, Bak, Bakkum, Deng, Dolan, Hoeschele, Welinder, Tillet, Pronin, Tillet, Dhariwal, Yuan, Dias, Lim, Arora, Troll, Lin, Lopes, Puri, Miyara, Leike, Gaubert, Zamani, Wang, Donnelly, Honsby, Smith, Sahai, Ramchandani, Huet, Carmichael, Zellers, Chen, Chen, Nigmatullin, Cheu, Jain, Altman, Schoenholz, Toizer, Miserendino, Agarwal, Culver, Ethersmith, Gray, Grove, Metzger, Hermani, Jain, Zhao, Wu, Jomoto, Wu, Shuaiqi, Xia, Phene, Papay, Narayanan, Coffey, Lee, Hall, Balaji, Broda, Stramer, Xu, Gogineni,
  Christianson, Sanders, Patwardhan, Cunninghman, Degry, Dimson, Raoux, Shadwell, Zheng, Underwood, Markov, Sherbakov, Rubin, Stasi, Kaftan, Heywood, Peterson, Walters, Eloundou, Qi, Moeller, Monaco, Kuo, Fomenko, Chang, Zheng, Zhou, Manassra, Sheu, Zaremba, Patil, Qian, Kim, Cheng, Zhang, He, Zhang, Jin, Dai, and Malkov}]{openai-2024-gpt4o-card}
OpenAI, :, Aaron Hurst, Adam Lerer, Adam~P. Goucher, Adam Perelman, Aditya Ramesh, Aidan Clark, AJ~Ostrow, Akila Welihinda, Alan Hayes, Alec Radford, Aleksander Mądry, Alex Baker-Whitcomb, Alex Beutel, Alex Borzunov, Alex Carney, Alex Chow, Alex Kirillov, Alex Nichol, Alex Paino, Alex Renzin, Alex~Tachard Passos, Alexander Kirillov, Alexi Christakis, Alexis Conneau, Ali Kamali, Allan Jabri, Allison Moyer, Allison Tam, Amadou Crookes, Amin Tootoochian, Amin Tootoonchian, Ananya Kumar, Andrea Vallone, Andrej Karpathy, Andrew Braunstein, Andrew Cann, Andrew Codispoti, Andrew Galu, Andrew Kondrich, Andrew Tulloch, Andrey Mishchenko, Angela Baek, Angela Jiang, Antoine Pelisse, Antonia Woodford, Anuj Gosalia, Arka Dhar, Ashley Pantuliano, Avi Nayak, Avital Oliver, Barret Zoph, Behrooz Ghorbani, Ben Leimberger, Ben Rossen, Ben Sokolowsky, Ben Wang, Benjamin Zweig, Beth Hoover, Blake Samic, Bob McGrew, Bobby Spero, Bogo Giertler, Bowen Cheng, Brad Lightcap, Brandon Walkin, Brendan Quinn, Brian Guarraci, Brian Hsu,
  Bright Kellogg, Brydon Eastman, Camillo Lugaresi, Carroll Wainwright, Cary Bassin, Cary Hudson, Casey Chu, Chad Nelson, Chak Li, Chan~Jun Shern, Channing Conger, Charlotte Barette, Chelsea Voss, Chen Ding, Cheng Lu, Chong Zhang, Chris Beaumont, Chris Hallacy, Chris Koch, Christian Gibson, Christina Kim, Christine Choi, Christine McLeavey, Christopher Hesse, Claudia Fischer, Clemens Winter, Coley Czarnecki, Colin Jarvis, Colin Wei, Constantin Koumouzelis, Dane Sherburn, Daniel Kappler, Daniel Levin, Daniel Levy, David Carr, David Farhi, David Mely, David Robinson, David Sasaki, Denny Jin, Dev Valladares, Dimitris Tsipras, Doug Li, Duc~Phong Nguyen, Duncan Findlay, Edede Oiwoh, Edmund Wong, Ehsan Asdar, Elizabeth Proehl, Elizabeth Yang, Eric Antonow, Eric Kramer, Eric Peterson, Eric Sigler, Eric Wallace, Eugene Brevdo, Evan Mays, Farzad Khorasani, Felipe~Petroski Such, Filippo Raso, Francis Zhang, Fred von Lohmann, Freddie Sulit, Gabriel Goh, Gene Oden, Geoff Salmon, Giulio Starace, Greg Brockman, Hadi
  Salman, Haiming Bao, Haitang Hu, Hannah Wong, Haoyu Wang, Heather Schmidt, Heather Whitney, Heewoo Jun, Hendrik Kirchner, Henrique~Ponde de~Oliveira~Pinto, Hongyu Ren, Huiwen Chang, Hyung~Won Chung, Ian Kivlichan, Ian O'Connell, Ian O'Connell, Ian Osband, Ian Silber, Ian Sohl, Ibrahim Okuyucu, Ikai Lan, Ilya Kostrikov, Ilya Sutskever, Ingmar Kanitscheider, Ishaan Gulrajani, Jacob Coxon, Jacob Menick, Jakub Pachocki, James Aung, James Betker, James Crooks, James Lennon, Jamie Kiros, Jan Leike, Jane Park, Jason Kwon, Jason Phang, Jason Teplitz, Jason Wei, Jason Wolfe, Jay Chen, Jeff Harris, Jenia Varavva, Jessica~Gan Lee, Jessica Shieh, Ji~Lin, Jiahui Yu, Jiayi Weng, Jie Tang, Jieqi Yu, Joanne Jang, Joaquin~Quinonero Candela, Joe Beutler, Joe Landers, Joel Parish, Johannes Heidecke, John Schulman, Jonathan Lachman, Jonathan McKay, Jonathan Uesato, Jonathan Ward, Jong~Wook Kim, Joost Huizinga, Jordan Sitkin, Jos Kraaijeveld, Josh Gross, Josh Kaplan, Josh Snyder, Joshua Achiam, Joy Jiao, Joyce Lee, Juntang
  Zhuang, Justyn Harriman, Kai Fricke, Kai Hayashi, Karan Singhal, Katy Shi, Kavin Karthik, Kayla Wood, Kendra Rimbach, Kenny Hsu, Kenny Nguyen, Keren Gu-Lemberg, Kevin Button, Kevin Liu, Kiel Howe, Krithika Muthukumar, Kyle Luther, Lama Ahmad, Larry Kai, Lauren Itow, Lauren Workman, Leher Pathak, Leo Chen, Li~Jing, Lia Guy, Liam Fedus, Liang Zhou, Lien Mamitsuka, Lilian Weng, Lindsay McCallum, Lindsey Held, Long Ouyang, Louis Feuvrier, Lu~Zhang, Lukas Kondraciuk, Lukasz Kaiser, Luke Hewitt, Luke Metz, Lyric Doshi, Mada Aflak, Maddie Simens, Madelaine Boyd, Madeleine Thompson, Marat Dukhan, Mark Chen, Mark Gray, Mark Hudnall, Marvin Zhang, Marwan Aljubeh, Mateusz Litwin, Matthew Zeng, Max Johnson, Maya Shetty, Mayank Gupta, Meghan Shah, Mehmet Yatbaz, Meng~Jia Yang, Mengchao Zhong, Mia Glaese, Mianna Chen, Michael Janner, Michael Lampe, Michael Petrov, Michael Wu, Michele Wang, Michelle Fradin, Michelle Pokrass, Miguel Castro, Miguel Oom~Temudo de~Castro, Mikhail Pavlov, Miles Brundage, Miles Wang, Minal
  Khan, Mira Murati, Mo~Bavarian, Molly Lin, Murat Yesildal, Nacho Soto, Natalia Gimelshein, Natalie Cone, Natalie Staudacher, Natalie Summers, Natan LaFontaine, Neil Chowdhury, Nick Ryder, Nick Stathas, Nick Turley, Nik Tezak, Niko Felix, Nithanth Kudige, Nitish Keskar, Noah Deutsch, Noel Bundick, Nora Puckett, Ofir Nachum, Ola Okelola, Oleg Boiko, Oleg Murk, Oliver Jaffe, Olivia Watkins, Olivier Godement, Owen Campbell-Moore, Patrick Chao, Paul McMillan, Pavel Belov, Peng Su, Peter Bak, Peter Bakkum, Peter Deng, Peter Dolan, Peter Hoeschele, Peter Welinder, Phil Tillet, Philip Pronin, Philippe Tillet, Prafulla Dhariwal, Qiming Yuan, Rachel Dias, Rachel Lim, Rahul Arora, Rajan Troll, Randall Lin, Rapha~Gontijo Lopes, Raul Puri, Reah Miyara, Reimar Leike, Renaud Gaubert, Reza Zamani, Ricky Wang, Rob Donnelly, Rob Honsby, Rocky Smith, Rohan Sahai, Rohit Ramchandani, Romain Huet, Rory Carmichael, Rowan Zellers, Roy Chen, Ruby Chen, Ruslan Nigmatullin, Ryan Cheu, Saachi Jain, Sam Altman, Sam Schoenholz, Sam
  Toizer, Samuel Miserendino, Sandhini Agarwal, Sara Culver, Scott Ethersmith, Scott Gray, Sean Grove, Sean Metzger, Shamez Hermani, Shantanu Jain, Shengjia Zhao, Sherwin Wu, Shino Jomoto, Shirong Wu, Shuaiqi, Xia, Sonia Phene, Spencer Papay, Srinivas Narayanan, Steve Coffey, Steve Lee, Stewart Hall, Suchir Balaji, Tal Broda, Tal Stramer, Tao Xu, Tarun Gogineni, Taya Christianson, Ted Sanders, Tejal Patwardhan, Thomas Cunninghman, Thomas Degry, Thomas Dimson, Thomas Raoux, Thomas Shadwell, Tianhao Zheng, Todd Underwood, Todor Markov, Toki Sherbakov, Tom Rubin, Tom Stasi, Tomer Kaftan, Tristan Heywood, Troy Peterson, Tyce Walters, Tyna Eloundou, Valerie Qi, Veit Moeller, Vinnie Monaco, Vishal Kuo, Vlad Fomenko, Wayne Chang, Weiyi Zheng, Wenda Zhou, Wesam Manassra, Will Sheu, Wojciech Zaremba, Yash Patil, Yilei Qian, Yongjik Kim, Youlong Cheng, Yu~Zhang, Yuchen He, Yuchen Zhang, Yujia Jin, Yunxing Dai, and Yury Malkov. 2024.
\newblock \href {http://arxiv.org/abs/2410.21276} {Gpt-4o system card}.

\bibitem[{OpenAI(2023)}]{openai-2024-gpt4-technical-report}
OpenAI. 2023.
\newblock \href {https://doi.org/10.48550/ARXIV.2303.08774} {{GPT-4} technical report}.
\newblock \emph{CoRR}, abs/2303.08774.

\bibitem[{Pan et~al.(2023)Pan, Wu, Lu, Luu, Wang, Kan, and Nakov}]{pan-etal-2023-fact}
Liangming Pan, Xiaobao Wu, Xinyuan Lu, Anh~Tuan Luu, William~Yang Wang, Min-Yen Kan, and Preslav Nakov. 2023.
\newblock \href {https://doi.org/10.18653/v1/2023.acl-long.386} {Fact-checking complex claims with program-guided reasoning}.
\newblock In \emph{Proceedings of the 61st Annual Meeting of the Association for Computational Linguistics (Volume 1: Long Papers)}, pages 6981--7004, Toronto, Canada. Association for Computational Linguistics.

\bibitem[{Paul(1993)}]{Paul1993}
Gabriele Paul. 1993.
\newblock \href {https://doi.org/10.1007/bf00849080} {Approaches to abductive reasoning: an overview}.
\newblock \emph{Artificial Intelligence Review}, 7(2):109–152.

\bibitem[{Plutynski(2011)}]{Plutynski2011}
Anya Plutynski. 2011.
\newblock \href {https://doi.org/10.1086/660746} {Four problems of abduction: A brief history}.
\newblock \emph{HOPOS: The Journal of the International Society for the History of Philosophy of Science}, 1(2):227–248.

\bibitem[{Pu et~al.(2021)Pu, Chung, Parikh, Gehrmann, and Sellam}]{pu2021learning}
Amy Pu, Hyung~Won Chung, Ankur~P Parikh, Sebastian Gehrmann, and Thibault Sellam. 2021.
\newblock Learning compact metrics for mt.
\newblock In \emph{Proceedings of EMNLP}.

\bibitem[{Radford et~al.(2019)Radford, Wu, Child, Luan, Amodei, and Sutskever}]{radford2019language}
Alec Radford, Jeff Wu, Rewon Child, David Luan, Dario Amodei, and Ilya Sutskever. 2019.
\newblock Language models are unsupervised multitask learners.

\bibitem[{Rawte et~al.(2023)Rawte, Chakraborty, Pathak, Sarkar, Tonmoy, Chadha, Sheth, and Das}]{rawte-etal-2023-troubling}
Vipula Rawte, Swagata Chakraborty, Agnibh Pathak, Anubhav Sarkar, S.M Towhidul~Islam Tonmoy, Aman Chadha, Amit Sheth, and Amitava Das. 2023.
\newblock \href {https://doi.org/10.18653/v1/2023.emnlp-main.155} {The troubling emergence of hallucination in large language models - an extensive definition, quantification, and prescriptive remediations}.
\newblock In \emph{Proceedings of the 2023 Conference on Empirical Methods in Natural Language Processing}, pages 2541--2573, Singapore. Association for Computational Linguistics.

\bibitem[{Saakyan et~al.(2021)Saakyan, Chakrabarty, and Muresan}]{saakyan-etal-2021-covid}
Arkadiy Saakyan, Tuhin Chakrabarty, and Smaranda Muresan. 2021.
\newblock \href {https://doi.org/10.18653/v1/2021.acl-long.165} {{COVID}-fact: Fact extraction and verification of real-world claims on {COVID}-19 pandemic}.
\newblock In \emph{Proceedings of the 59th Annual Meeting of the Association for Computational Linguistics and the 11th International Joint Conference on Natural Language Processing (Volume 1: Long Papers)}, pages 2116--2129, Online. Association for Computational Linguistics.

\bibitem[{Saparov et~al.(2023)Saparov, Pang, Padmakumar, Joshi, Kazemi, Kim, and He}]{saparov2023testing}
Abulhair Saparov, Richard~Yuanzhe Pang, Vishakh Padmakumar, Nitish Joshi, Seyed~Mehran Kazemi, Najoung Kim, and He~He. 2023.
\newblock \href {http://arxiv.org/abs/2305.15269} {Testing the general deductive reasoning capacity of large language models using ood examples}.

\bibitem[{Schaeffer et~al.(2023)Schaeffer, Miranda, and Koyejo}]{schaeffer2023emergent}
Rylan Schaeffer, Brando Miranda, and Sanmi Koyejo. 2023.
\newblock \href {http://arxiv.org/abs/2304.15004} {Are emergent abilities of large language models a mirage?}

\bibitem[{Schlichtkrull et~al.(2023)Schlichtkrull, Ousidhoum, and Vlachos}]{schlichtkrull-etal-2023-intended}
Michael Schlichtkrull, Nedjma Ousidhoum, and Andreas Vlachos. 2023.
\newblock \href {https://doi.org/10.18653/v1/2023.findings-emnlp.577} {The intended uses of automated fact-checking artefacts: Why, how and who}.
\newblock In \emph{Findings of the Association for Computational Linguistics: EMNLP 2023}, pages 8618--8642, Singapore. Association for Computational Linguistics.

\bibitem[{Schuster et~al.(2021)Schuster, Fisch, and Barzilay}]{schuster-etal-2021-get}
Tal Schuster, Adam Fisch, and Regina Barzilay. 2021.
\newblock \href {https://www.aclweb.org/anthology/2021.naacl-main.52} {Get your vitamin {C}! robust fact verification with contrastive evidence}.
\newblock In \emph{Proceedings of the 2021 Conference of the North American Chapter of the Association for Computational Linguistics: Human Language Technologies}, pages 624--643, Online. Association for Computational Linguistics.

\bibitem[{Shapira et~al.(2023)Shapira, Levy, Alavi, Zhou, Choi, Goldberg, Sap, and Shwartz}]{Shapira2023CleverHO}
Natalie Shapira, Mosh Levy, Seyed~Hossein Alavi, Xuhui Zhou, Yejin Choi, Yoav Goldberg, Maarten Sap, and Vered Shwartz. 2023.
\newblock \href {https://api.semanticscholar.org/CorpusID:258865502} {Clever hans or neural theory of mind? stress testing social reasoning in large language models}.
\newblock \emph{ArXiv}, abs/2305.14763.

\bibitem[{Shi et~al.(2023)Shi, Xue, Wang, Zhou, Zhang, Zhou, Tan, and Mei}]{shi2023language}
Xiaoming Shi, Siqiao Xue, Kangrui Wang, Fan Zhou, James~Y. Zhang, Jun Zhou, Chenhao Tan, and Hongyuan Mei. 2023.
\newblock \href {http://arxiv.org/abs/2305.16646} {Language models can improve event prediction by few-shot abductive reasoning}.

\bibitem[{Sileo and Lernould(2023)}]{sileo-lernould-2023-mindgames}
Damien Sileo and Antoine Lernould. 2023.
\newblock \href {https://doi.org/10.18653/v1/2023.findings-emnlp.303} {{M}ind{G}ames: Targeting theory of mind in large language models with dynamic epistemic modal logic}.
\newblock In \emph{Findings of the Association for Computational Linguistics: EMNLP 2023}, pages 4570--4577, Singapore. Association for Computational Linguistics.

\bibitem[{Song et~al.(2024)Song, Chim, Tsakalidis, Ive, Atzil-Slonim, and Liakata}]{song-etal-2024-combining}
Jiayu Song, Jenny Chim, Adam Tsakalidis, Julia Ive, Dana Atzil-Slonim, and Maria Liakata. 2024.
\newblock \href {https://doi.org/10.18653/v1/2024.findings-acl.873} {Combining hierachical {VAE}s with {LLM}s for clinically meaningful timeline summarisation in social media}.
\newblock In \emph{Findings of the Association for Computational Linguistics: ACL 2024}, pages 14651--14672, Bangkok, Thailand. Association for Computational Linguistics.

\bibitem[{Sourati et~al.(2024)Sourati, Ilievski, Sommerauer, and Jiang}]{sourati-etal-2024-arn}
Zhivar Sourati, Filip Ilievski, Pia Sommerauer, and Yifan Jiang. 2024.
\newblock \href {https://doi.org/10.1162/tacl_a_00688} {{ARN}: Analogical reasoning on narratives}.
\newblock \emph{Transactions of the Association for Computational Linguistics}, 12:1063--1086.

\bibitem[{Sprague et~al.(2024)Sprague, Yin, Rodriguez, Jiang, Wadhwa, Singhal, Zhao, Ye, Mahowald, and Durrett}]{sprague2024cotcotchainofthoughthelps}
Zayne Sprague, Fangcong Yin, Juan~Diego Rodriguez, Dongwei Jiang, Manya Wadhwa, Prasann Singhal, Xinyu Zhao, Xi~Ye, Kyle Mahowald, and Greg Durrett. 2024.
\newblock \href {http://arxiv.org/abs/2409.12183} {To cot or not to cot? chain-of-thought helps mainly on math and symbolic reasoning}.

\bibitem[{Stevenson et~al.(2024)Stevenson, Pafford, van~der Maas, and Mitchell}]{stevenson-etal-2024-large}
Claire~E. Stevenson, Alexandra Pafford, Han L.~J. van~der Maas, and Melanie Mitchell. 2024.
\newblock \href {https://doi.org/10.48550/ARXIV.2411.02348} {Can large language models generalize analogy solving like people can?}
\newblock \emph{CoRR}, abs/2411.02348.

\bibitem[{Strong et~al.(2024)Strong, Aly, and Vlachos}]{strong-etal-2024-zero}
Marek Strong, Rami Aly, and Andreas Vlachos. 2024.
\newblock \href {https://doi.org/10.18653/v1/2024.findings-emnlp.991} {Zero-shot fact verification via natural logic and large language models}.
\newblock In \emph{Findings of the Association for Computational Linguistics: EMNLP 2024}, pages 17021--17035, Miami, Florida, USA. Association for Computational Linguistics.

\bibitem[{Tan et~al.(2024)Tan, Desai, and Sengamedu}]{tan-etal-2024-enhancing-fact}
Fiona~Anting Tan, Jay Desai, and Srinivasan~H. Sengamedu. 2024.
\newblock \href {https://doi.org/10.18653/v1/2024.fever-1.20} {Enhancing fact verification with causal knowledge graphs and transformer-based retrieval for deductive reasoning}.
\newblock In \emph{Proceedings of the Seventh Fact Extraction and VERification Workshop (FEVER)}, pages 151--169, Miami, Florida, USA. Association for Computational Linguistics.

\bibitem[{Tang et~al.(2024)Tang, Laban, and Durrett}]{tang-etal-2024-minicheck}
Liyan Tang, Philippe Laban, and Greg Durrett. 2024.
\newblock \href {https://doi.org/10.18653/v1/2024.emnlp-main.499} {{M}ini{C}heck: Efficient fact-checking of {LLM}s on grounding documents}.
\newblock In \emph{Proceedings of the 2024 Conference on Empirical Methods in Natural Language Processing}, pages 8818--8847, Miami, Florida, USA. Association for Computational Linguistics.

\bibitem[{Terwiesch(2023)}]{Terwiesch2023}
Christian Terwiesch. 2023.
\newblock Would chat gpt get a wharton mba? a prediction based on its performance in the operations management course.
\newblock Technical report, Mack Institute for Innovation Management at the Wharton School, University of Pennsylvania.

\bibitem[{Todd et~al.(2023)Todd, Li, Sharma, Mueller, Wallace, and Bau}]{Todd2023FunctionVI}
Eric Todd, Millicent Li, Arnab~Sen Sharma, Aaron Mueller, Byron~C. Wallace, and David Bau. 2023.
\newblock \href {https://api.semanticscholar.org/CorpusID:264439657} {Function vectors in large language models}.
\newblock \emph{ArXiv}, abs/2310.15213.

\bibitem[{Treutlein et~al.(2024)Treutlein, Choi, Betley, Marks, Anil, Grosse, and Evans}]{treutlein-etal-2024-connecting}
Johannes Treutlein, Dami Choi, Jan Betley, Samuel Marks, Cem Anil, Roger~Baker Grosse, and Owain Evans. 2024.
\newblock \href {https://openreview.net/forum?id=7FokMz6U8n} {Connecting the dots: {LLM}s can infer and verbalize latent structure from disparate training data}.
\newblock In \emph{The Thirty-eighth Annual Conference on Neural Information Processing Systems}.

\bibitem[{Ullman(2023)}]{ullman2023large}
Tomer Ullman. 2023.
\newblock \href {http://arxiv.org/abs/2302.08399} {Large language models fail on trivial alterations to theory-of-mind tasks}.

\bibitem[{Wang and Shu(2023)}]{wang-shu-2023-explainable}
Haoran Wang and Kai Shu. 2023.
\newblock \href {https://doi.org/10.18653/v1/2023.findings-emnlp.416} {Explainable claim verification via knowledge-grounded reasoning with large language models}.
\newblock In \emph{Findings of the Association for Computational Linguistics: EMNLP 2023}, pages 6288--6304, Singapore. Association for Computational Linguistics.

\bibitem[{Wason and Johnson{-}Laird(1972)}]{Wason1972-WASPOR-3}
Peter~Cathcart Wason and Philip~Nicholas Johnson{-}Laird. 1972.
\newblock \emph{Psychology of Reasoning: Structure and Content}.
\newblock Harvard University Press, Cambridge, MA, USA.

\bibitem[{Webb et~al.(2023)Webb, Holyoak, and Lu}]{Webb2023}
Taylor Webb, Keith~J. Holyoak, and Hongjing Lu. 2023.
\newblock \href {https://doi.org/10.1038/s41562-023-01659-w} {Emergent analogical reasoning in large language models}.
\newblock \emph{Nature Human Behaviour}, 7(9):1526–1541.

\bibitem[{Wei et~al.(2022)Wei, Tay, Bommasani, Raffel, Zoph, Borgeaud, Yogatama, Bosma, Zhou, Metzler, Chi, Hashimoto, Vinyals, Liang, Dean, and Fedus}]{wei2022emergent}
Jason Wei, Yi~Tay, Rishi Bommasani, Colin Raffel, Barret Zoph, Sebastian Borgeaud, Dani Yogatama, Maarten Bosma, Denny Zhou, Donald Metzler, Ed~H. Chi, Tatsunori Hashimoto, Oriol Vinyals, Percy Liang, Jeff Dean, and William Fedus. 2022.
\newblock \href {http://arxiv.org/abs/2206.07682} {Emergent abilities of large language models}.

\bibitem[{Wei et~al.(2023)Wei, Wang, Schuurmans, Bosma, Ichter, Xia, Chi, Le, and Zhou}]{wei2023chainofthought}
Jason Wei, Xuezhi Wang, Dale Schuurmans, Maarten Bosma, Brian Ichter, Fei Xia, Ed~Chi, Quoc Le, and Denny Zhou. 2023.
\newblock \href {http://arxiv.org/abs/2201.11903} {Chain-of-thought prompting elicits reasoning in large language models}.

\bibitem[{Wu et~al.(2023)Wu, Qiu, Ross, Akyürek, Chen, Wang, Kim, Andreas, and Kim}]{wu2023reasoning}
Zhaofeng Wu, Linlu Qiu, Alexis Ross, Ekin Akyürek, Boyuan Chen, Bailin Wang, Najoung Kim, Jacob Andreas, and Yoon Kim. 2023.
\newblock \href {http://arxiv.org/abs/2307.02477} {Reasoning or reciting? exploring the capabilities and limitations of language models through counterfactual tasks}.

\bibitem[{Xu et~al.(2025)Xu, Lin, Han, Zhao, Liu, and Cambria}]{10.1109/TKDE.2025.3536008}
Fangzhi Xu, Qika Lin, Jiawei Han, Tianzhe Zhao, Jun Liu, and Erik Cambria. 2025.
\newblock \href {https://doi.org/10.1109/TKDE.2025.3536008} {Are large language models really good logical reasoners? a comprehensive evaluation and beyond}.
\newblock \emph{IEEE Trans. on Knowl. and Data Eng.}, 37(4):1620–1634.

\bibitem[{Xu et~al.(2023)Xu, Wang, Li, Luo, Wang, Liu, and Liu}]{Xu2023ExploringLL}
Yuzhuang Xu, Shuo Wang, Peng Li, Fuwen Luo, Xiaolong Wang, Weidong Liu, and Yang Liu. 2023.
\newblock \href {https://api.semanticscholar.org/CorpusID:261681932} {Exploring large language models for communication games: An empirical study on werewolf}.
\newblock \emph{ArXiv}, abs/2309.04658.

\bibitem[{Yang et~al.(2024)Yang, Gribovskaya, Kassner, Geva, and Riedel}]{yang-etal-2024-large-language-models}
Sohee Yang, Elena Gribovskaya, Nora Kassner, Mor Geva, and Sebastian Riedel. 2024.
\newblock \href {https://doi.org/10.18653/v1/2024.acl-long.550} {Do large language models latently perform multi-hop reasoning?}
\newblock In \emph{Proceedings of the 62nd Annual Meeting of the Association for Computational Linguistics (Volume 1: Long Papers)}, pages 10210--10229, Bangkok, Thailand. Association for Computational Linguistics.

\bibitem[{Yasunaga et~al.(2024)Yasunaga, Chen, Li, Pasupat, Leskovec, Liang, Chi, and Zhou}]{yasunaga-etal-2024-large}
Michihiro Yasunaga, Xinyun Chen, Yujia Li, Panupong Pasupat, Jure Leskovec, Percy Liang, Ed~H. Chi, and Denny Zhou. 2024.
\newblock \href {https://openreview.net/forum?id=AgDICX1h50} {Large language models as analogical reasoners}.
\newblock In \emph{The Twelfth International Conference on Learning Representations, {ICLR} 2024, Vienna, Austria, May 7-11, 2024}. OpenReview.net.

\bibitem[{Ye et~al.(2023)Ye, Kuribayashi, Suzuki, Kobayashi, and Funayama}]{ye-etal-2023-assessing}
Mengyu Ye, Tatsuki Kuribayashi, Jun Suzuki, Goro Kobayashi, and Hiroaki Funayama. 2023.
\newblock \href {https://doi.org/10.18653/v1/2023.emnlp-main.912} {Assessing step-by-step reasoning against lexical negation: A case study on syllogism}.
\newblock In \emph{Proceedings of the 2023 Conference on Empirical Methods in Natural Language Processing}, pages 14753--14773, Singapore. Association for Computational Linguistics.

\bibitem[{Ye et~al.(2024)Ye, Wang, Choi, Lu, Sharma, Shen, Tiyyala, Andrews, and Khashabi}]{ye-etal-2024-analobench}
Xiao Ye, Andrew Wang, Jacob Choi, Yining Lu, Shreya Sharma, Lingfeng Shen, Vijay~Murari Tiyyala, Nicholas Andrews, and Daniel Khashabi. 2024.
\newblock \href {https://doi.org/10.18653/v1/2024.emnlp-main.725} {{A}nalo{B}ench: Benchmarking the identification of abstract and long-context analogies}.
\newblock In \emph{Proceedings of the 2024 Conference on Empirical Methods in Natural Language Processing}, pages 13060--13082, Miami, Florida, USA. Association for Computational Linguistics.

\bibitem[{Yu et~al.(2024)Yu, He, and Ying}]{yu-etal-2024-thought}
Junchi Yu, Ran He, and Zhitao Ying. 2024.
\newblock \href {https://openreview.net/forum?id=SBoRhRCzM3} {Thought propagation: an analogical approach to complex reasoning with large language models}.
\newblock In \emph{The Twelfth International Conference on Learning Representations, {ICLR} 2024, Vienna, Austria, May 7-11, 2024}. OpenReview.net.

\bibitem[{Yu et~al.(2023)Yu, Zhang, Yu, and Jiang}]{yu-etal-2023-pre}
Mengxia Yu, Zhihan Zhang, Wenhao Yu, and Meng Jiang. 2023.
\newblock \href {https://doi.org/10.18653/v1/2023.emnlp-main.763} {Pre-training language models for comparative reasoning}.
\newblock In \emph{Proceedings of the 2023 Conference on Empirical Methods in Natural Language Processing}, pages 12421--12433, Singapore. Association for Computational Linguistics.

\bibitem[{Yuan et~al.(2021)Yuan, Neubig, and Liu}]{NEURIPS2021_e4d2b6e6}
Weizhe Yuan, Graham Neubig, and Pengfei Liu. 2021.
\newblock \href {https://proceedings.neurips.cc/paper/2021/file/e4d2b6e6fdeca3e60e0f1a62fee3d9dd-Paper.pdf} {Bartscore: Evaluating generated text as text generation}.
\newblock In \emph{Advances in Neural Information Processing Systems}, volume~34, pages 27263--27277. Curran Associates, Inc.

\bibitem[{Zhang et~al.(2024)Zhang, Tong, Zhang, and Zhang}]{zhang-etal-2024-probing}
Chenyang Zhang, Haibo Tong, Bin Zhang, and Dongyu Zhang. 2024.
\newblock \href {https://doi.org/10.48550/ARXIV.2408.14380} {Probing causality manipulation of large language models}.
\newblock \emph{CoRR}, abs/2408.14380.

\bibitem[{Zhang et~al.(2020)Zhang, Kishore, Wu, Weinberger, and Artzi}]{bert-score}
Tianyi Zhang, Varsha Kishore, Felix Wu, Kilian~Q. Weinberger, and Yoav Artzi. 2020.
\newblock \href {https://openreview.net/forum?id=SkeHuCVFDr} {Bertscore: Evaluating text generation with bert}.
\newblock In \emph{International Conference on Learning Representations}.

\bibitem[{Zhao et~al.(2023)Zhao, Chiu, Cardie, and Rush}]{zhao-etal-2023-abductive}
Wenting Zhao, Justin Chiu, Claire Cardie, and Alexander Rush. 2023.
\newblock \href {https://doi.org/10.18653/v1/2023.acl-long.831} {Abductive commonsense reasoning exploiting mutually exclusive explanations}.
\newblock In \emph{Proceedings of the 61st Annual Meeting of the Association for Computational Linguistics (Volume 1: Long Papers)}, pages 14883--14896, Toronto, Canada. Association for Computational Linguistics.

\bibitem[{Zubiaga et~al.(2018)Zubiaga, Aker, Bontcheva, Liakata, and Procter}]{zubiaga-etal-2019-detection}
Arkaitz Zubiaga, Ahmet Aker, Kalina Bontcheva, Maria Liakata, and Rob Procter. 2018.
\newblock \href {https://doi.org/10.1145/3161603} {Detection and resolution of rumours in social media: A survey}.
\newblock \emph{ACM Comput. Surv.}, 51(2).

\bibitem[{Zubiaga et~al.(2016)Zubiaga, Hoi, Liakata, and Procter}]{pheme}
Arkaitz Zubiaga, Geraldine Wong~Sak Hoi, Maria Liakata, and Rob Procter. 2016.
\newblock \href {https://wrap.warwick.ac.uk/134772/} {Pheme dataset of rumours and non-rumours}.

\end{thebibliography}
